\documentclass{article}

\usepackage[utf8]{inputenc}
\usepackage[margin=1in]{geometry}
\usepackage[numbers,compress]{natbib}
\usepackage[T1]{fontenc}
\usepackage{hyperref}
\usepackage{url}
\usepackage{graphicx}
\usepackage{microtype}
\usepackage{nicefrac}
\usepackage{amsfonts}

\AtBeginDocument{%
  }

\usepackage{xcolor}
\usepackage{wrapfig}

\usepackage{nicefrac}
\usepackage{hyperref}
\hypersetup{
    colorlinks=true,
    linkcolor=blue,
    filecolor=magenta,      
    urlcolor=cyan
}
\usepackage{mathtools}
\usepackage{amsmath,amsfonts}

\usepackage{amssymb}
\usepackage[ruled,vlined,linesnumbered]{algorithm2e}
\usepackage{algorithmic}
\usepackage{makecell}
\usepackage{diagbox}
\usepackage{multirow}
\usepackage{textgreek}
\usepackage{comment}

\usepackage{caption}
\usepackage{subcaption}

\usepackage{listings}

\usepackage{amsthm}

\usepackage[nameinlink,capitalize]{cleveref}

\crefname{section}{\S}{\S}
\Crefname{section}{\S}{\S}
\crefname{appendix}{App.}{Apps.}
\Crefname{appendix}{App.}{Apps.}
\crefname{theorem}{Thm.}{Thms.}
\Crefname{theorem}{Thm.}{Thms.}
\crefname{proposition}{Prop.}{Props.}
\Crefname{proposition}{Prop.}{Props.}
\crefname{algorithm}{Alg.}{Algs.}
\Crefname{algorithm}{Alg.}{Algs.}
\crefname{assumption}{Asm.}{Asms.}
\Crefname{assumption}{Asm.}{Asms.}
\crefname{mechanism}{Mech.}{Mechs.}
\Crefname{mechanism}{Mech.}{Mechs.}

\usepackage{mfirstuc}
\usepackage[T1]{fontenc}

\newcounter{packednmbr}

\definecolor{darkgreen}{RGB}{0,140,0}

\newcommand{\red}[1]{\textcolor{red}{#1}}

\newcommand{\green}[1]{\textcolor{darkgreen}{\textbf{#1}}}

\newcommand{\revise}[1]{#1}
\newcommand{\orange}[1]{\textcolor{orange}{#1}}

\newcommand{\bra}[1]{\left( #1 \right)}

\newcommand{\brc}[1]{\left\{ #1 \right\}}

\newcommand{\precision}{KNN-Precision}
\newcommand{\recall}{KNN-Recall}
\newcommand{\fid}{FID}

\newcommand{\base}{\mathcal{D}}

\newcommand{\sample}{D}
\newcommand{\instruction}{R}
\newcommand{\tool}{F}
\newcommand{\finaloutput}{O}

\newcommand{\toolusage}{f}
\newcommand{\instructionround}{r}
\newcommand{\functionset}{\mathcal{F}}

\newcommand{\KND}{KND}
\newcommand{\AM}{AM}
\newcommand{\entropy}{AD}

\newcommand{\name}{\texttt{SynAE}}

\newcommand{\tooluse}{TUM}
\newcommand{\toolnumber}{TCNM}
\newcommand{\toolplanning}{k-Step Planning}
\newcommand{\toolplanningtwo}{2-Step Planning}

\newcommand{\successrate}{VR}

\newcommand{\difficulty}{TDD}
\newcommand{\ranking}{RD}

\newcommand{\blankfilling}{p}
\newcommand{\oversampling}{r}
\newcommand{\ice}{k}
\newcommand{\invalid}{v}

\newcommand{\dis}{\text{Dis}}
\newcommand{\distribution}{\omega}

\newcommand{\smatrix}{\mathbf{K}}

\usepackage{booktabs}
\usepackage{longtable}
\usepackage{multirow}
\usepackage{caption}
\usepackage{subcaption}
\usepackage{rotating}

\usepackage{amsmath}
\usepackage{amssymb}
\usepackage{mathtools}
\usepackage{amsthm}

\usepackage{todonotes}
\usepackage{enumitem}

\theoremstyle{plain}

\theoremstyle{definition}

\theoremstyle{remark}

\usepackage{tcolorbox}

\usepackage{setspace}

\crefformat{section}{§#2#1#3}

\newcommand{\camerareadydelete}[1]{}

\title{\name{}: A Framework for Measuring the Quality of Synthetic Data for Tool-Calling Agent Evaluations}

\author{
  Shuaiqi Wang \\
  \small{Carnegie Mellon University}\\
  \small{\texttt{shuaiqiw@andrew.cmu.edu}}
  \and
  Aadyaa Maddi \\
  \small{Carnegie Mellon University}\\
  \small{\texttt{amaddi@andrew.cmu.edu}}
  \and
  Zinan Lin \\
  \small{Microsoft Research}\\
  \small{\texttt{zinanlin@microsoft.com}}
  \and
  Giulia Fanti \\
  \small{Carnegie Mellon University}\\
  \small{\texttt{gfanti@andrew.cmu.edu}}
}

\date{}

\begin{document}

\maketitle

\begin{abstract}
Today, tool-calling agents are commonly evaluated or tested on static datasets of \emph{execution traces}, including input commands, agent responses, and associated tool calls. 
However, internal production datasets are often insufficient or unusable for testing; for example, they may contain sensitive or proprietary data, or they may be too sparse to support comprehensive testing (especially pre-deployment). 
In these settings, practitioners are increasingly replacing or augmenting real datasets with synthetic ones for evaluation purposes. 
A key challenge is quantifying the relation between these synthetic datasets and the real data.
We introduce \name{}, an evaluation framework for assessing how well synthetic benchmarks for multi-turn, tool-calling agents replicate and augment the characteristics of real data trajectories. \name{} assesses the validity, fidelity, and diversity of synthetic data across four metric categories: (i) task instructions and intermediate responses, (ii) tool calls, (iii) final outputs, and (iv) downstream evaluation. We evaluate \name{} using recent agent benchmarks and test common synthetic data failure modes via realistic and controlled generation schemes.
\name{} detects fine-grained variations in data validity, fidelity and diversity, and shows that no single metric is sufficient to fully characterize synthetic data quality, motivating a multi-axis evaluation of synthetic data for agent testing. \revise{A demo of \name{} is available at \url{https://synae-2026-synae-demo.static.hf.space/index.html}, with code at \url{https://github.com/wsqwsq/SynAE}.}
\end{abstract}
\section{Introduction}

Agent evaluation and testing is a nascent but critical component of pre-deployment processes for production agentic workflows \citep{pan2025measuring, anthropic2026demystifying}. 
Today, {tool-calling} agent evaluations are often (but not always) run on static \emph{baseline datasets}\footnote{These evaluation datasets are also commonly referred to as \emph{benchmarks}; we use both terms interchangeably in this paper.} consisting of a trace generated from agent interactions; these typically include user inputs, tool calls, intermediate interactions with the agent, and a final output \citep{yehudai2025survey, mohammadi2025evaluation}.
Such datasets are often collected from real-user interactions with an environment and/or synthesized by measuring scripted interactions with a dynamic environment.
While the design of proper agent evaluations is its own active research area \citep{kapoor2025holistic, alonso2025evaluating, zhu2025establishing}, common evaluation metrics measure whether the agent selects the correct tools and produces the desired final output \citep{yehudai2025survey, mohammadi2025evaluation, anthropic2026demystifying}.

\begin{figure*}[htbp]
\vspace{-2.5mm}
    \centering
    \begin{subfigure}{0.48\textwidth}
        \centering
        \includegraphics[width=\textwidth]{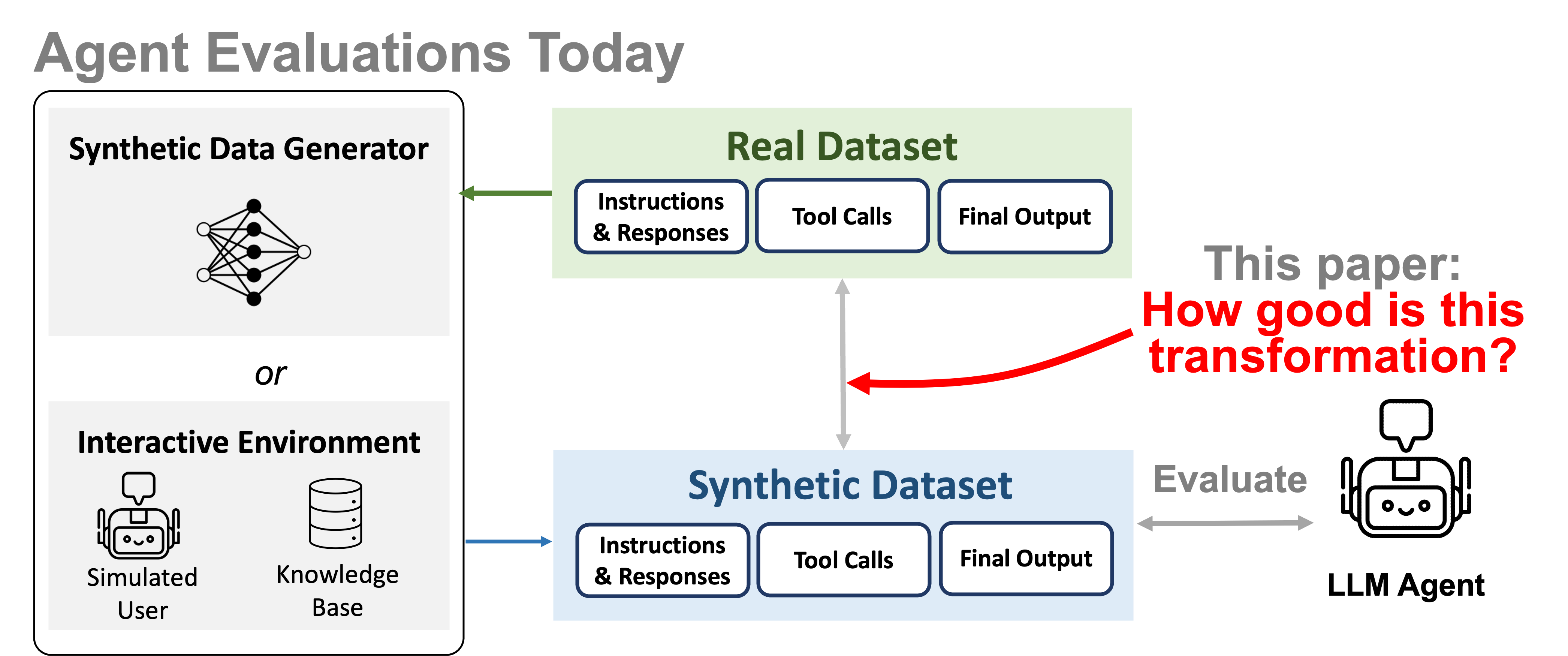}
        \caption{Today, agents are commonly evaluated on synthetic datasets modeled after a real dataset of agent trajectories, e.g., for privacy or augmentation reasons.
        However, it is often unclear whether these synthetic datasets are representative of the real dataset.
        Our evaluation framework, \name{}, measures the semantic and structural \emph{similarity} between real and synthetic datasets of execution trajectories, and the \emph{validity} and \emph{diversity} of the synthetic data.}
        \label{fig:scenario}
    \end{subfigure}
    \hfill
    \begin{subfigure}{0.5\textwidth}
        \centering
        \includegraphics[width=\textwidth]{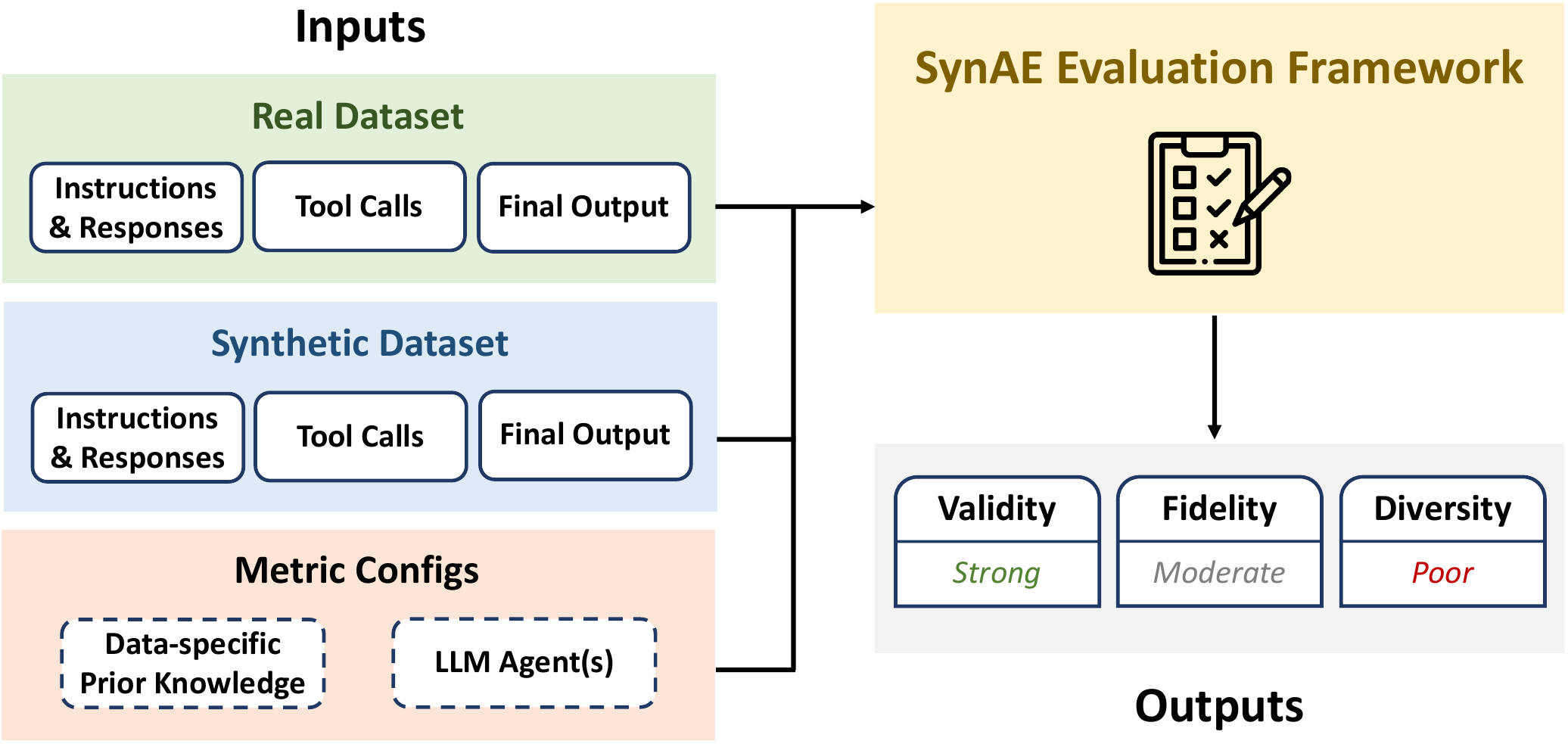}
        \caption{The \name{} evaluation pipeline takes as input real and synthetic datasets (task completion trajectories), along with optional metric configurations that may specify data-specific prior knowledge and one or more LLM agents. 
        The LLM agents are used only to compare downstream performance between real and synthetic input datasets.
        The framework evaluates synthetic data along three dimensions: validity, fidelity, and diversity.
        }
        \label{fig:pipeline}
    \end{subfigure}
    \vspace{-0.5mm}
    \caption{The \name{} framework  evaluates the quality of synthetic data used in agent evaluations. 
    }
    \label{fig:main}
    \vspace{-3mm}
\end{figure*}

In many practical situations, existing baseline datasets cannot be directly used or are insufficient for agent evaluation, 
either because they contain sensitive user data (e.g., emails, travel details) subject to privacy restrictions
\citep{tamkin2024clio,capitalone} or because they are too small for comprehensive testing \citep{pan2025measuring}.
Practitioners therefore increasingly use \emph{synthetic datasets} as replacements for or augmentations to real execution traces in evaluation pipelines (\Cref{fig:scenario}) \citep{qin2023toolllm,tang2023toolalpaca,iskander2024quality}. Synthetic data may be generated by directly synthesizing trajectories or by synthesizing inputs to an interactive environment.

Although synthetic datasets are increasingly used for agent evaluation, typical workflows lack systematic quality checks against the real baseline data. 
Indeed, \textbf{the current literature on testing agents with synthetic data provides almost no quantitative methods for evaluating the quality of such synthetic data,} %
leaving operators with little visibility into evaluation gaps.

In this work, we develop a comprehensive evaluation framework, \name{}, to assess how well synthetic trajectories \emph{replicate} and \emph{augment} the characteristics of real data trajectories, including task instructions and responses, and the associated reference tool calls and outputs. 
As illustrated in \Cref{fig:pipeline},  \name{}  takes as input a real baseline dataset, a synthetic dataset, and optional metric configurations that may specify data-specific prior knowledge and one or more LLM agents; the input agent(s) need not be the same as the one being evaluated---each agent input to \name{}  is used purely to compare downstream performance between real and synthetic input datasets.

\name{} evaluates synthetic data along three properties: (1) validity, (2) fidelity, and (3) diversity. 
For \emph{validity}, we evaluate whether synthetic tool calls and outputs successfully fulfill the given instructions, using LLM-as-a-judge by default or rule-based checkers when available.
We measure \emph{fidelity} by assessing how similar the real and the synthetic data are; specifically, we compute metrics on both real and synthetic data and evaluate their similarity. 
We  quantify the \emph{diversity} of synthetic data with entropy-based metrics for various representations of the dataset.
For each property, \name{} computes a suite of metrics to evaluate (a subset of): (1) task instructions and responses, (2) tool calls, (3) final outputs, and (4) downstream evaluation.

\revise{We demonstrate the utility of \name{} with three recent agent benchmarks playing the role of the ``real dataset'' in \cref{fig:scenario}: T1 \citep{chakraborty2025t1}, BFCL \citep{patilberkeley}, and ACP \citep{kokel2025acpbench}. We then construct synthetic datasets using NVIDIA NeMo \cite{nvidia_nemo}, an industry-standard synthetic data tool, as well as custom synthetic data generation methods designed to simulate common real-world pitfalls, such as degraded data fidelity and limited diversity.} Our experiments show that \name{} captures fine-grained variations in the validity, fidelity, and diversity of synthetic task-completion trajectories. They further suggest that no single metric is sufficient to fully characterize synthetic data quality, underscoring the need for \name{}.
Overall, we view \name{} as a plug-in component for agentic workflows, allowing operators to automatically evaluate the quality of synthetic benchmark datasets.

\subsection{Related Work}

We provide a brief overview of related works, and provide a detailed discussion in \cref{app:related_work}.

Several prior works evaluate synthetic benchmarks for LLMs using factors such as task difficulty, realism, and whether model rankings are preserved \cite{gill2025has,maheshwari2024efficacy,xiong2025probe,majurski2025grounding}. However, these works mainly focus on standard NLP settings rather than agent benchmarks, where evaluation is more difficult due to multi-step decision making, interaction with tools, and the need to attribute failures to different parts of the agent pipeline. Recent evaluations on synthetic benchmarks for tool-calling agents has largely focused on single-turn settings or instruction-level quality \cite{shen2024taskbench,iskander2024quality,zhu2025establishing,alonso2025evaluating,pan2025measuring}, without systematically evaluating associated reference tool calls, outputs, or multi-step dependencies.

Multi-turn tool-calling agents are commonly evaluated using end-to-end task success on state-based checkers \citep{yao2024tau,liu2023agentbench,zhou2023webarena,yang2023intercode,xie2024osworld,jimenez2023swe}, response-based criteria \citep{patilberkeley,li2025toolrm,zhanggecko}, or LLM-as-a-judge \citep{qin2023toolllm,pan2024autonomous,xue2025illusion,lu2025agentrewardbench}. These methods evaluate agent performance on a given benchmark, whereas our goal is to evaluate the benchmark itself by providing quantitative metrics for validity, fidelity, and diversity across instructions and responses, tool calls, final outputs, and downstream evaluation.

\section{\name{} Framework}

\vspace{-0.5mm}
The \name{} framework assesses how well synthetic agent trajectories replicate and augment the characteristics of a real dataset. 
It  quantifies the \emph{validity}, \emph{fidelity}, and \emph{diversity} of synthetic data.

\begin{figure}[htbp]
\vspace{-0.5mm}
    \centering
    \includegraphics[width=0.73\linewidth]{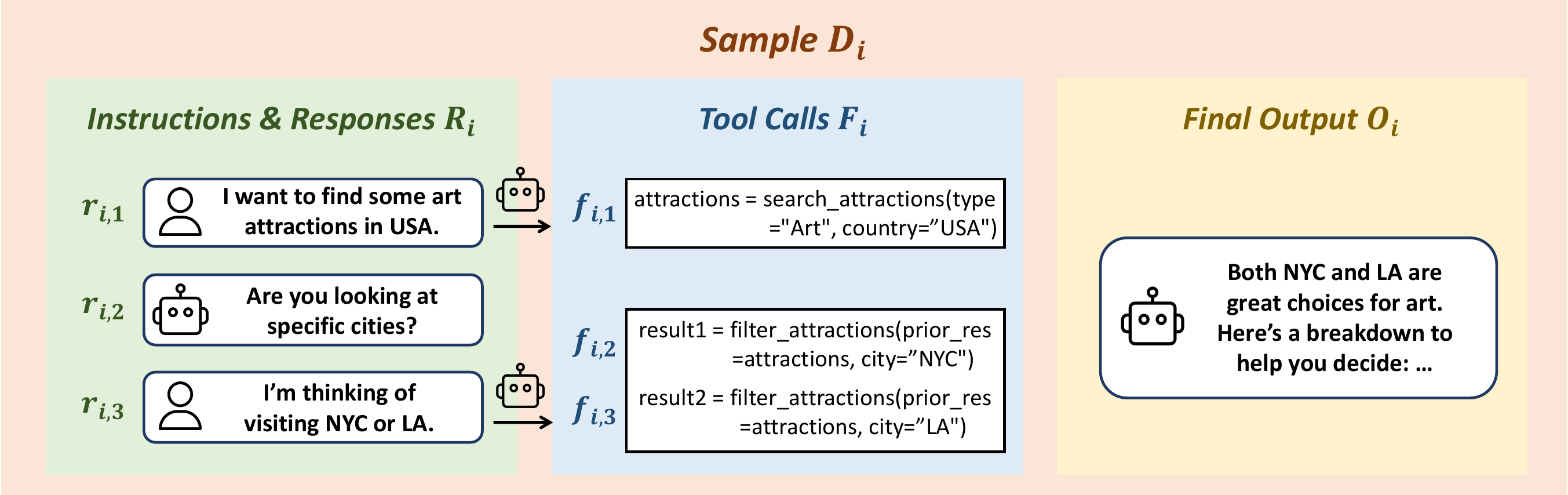}
    \vspace{-0.5mm}
    \caption{Agent trajectory from T1 \cite{chakraborty2025t1} benchmark dataset, with notation for each component. %
    }
    \label{fig:data_illustration}
    \vspace{-2.5mm}
\end{figure}

\paragraph{\textbf{Notation and setup}} 
Consider a dataset $\base = \brc{\sample_i}_{i=1}^{m}$ of $m$ samples (or agent trajectories).
\cref{fig:data_illustration} illustrates a single sample trajectory $\sample_i$, which consists of a set of instructions and responses $\instruction_i$, tool calls $\tool_i$, and a textual output $\finaloutput_i$. $\instruction_i$ is a sequence containing multiple instructions and responses $\instruction_i=(\instructionround_{i,1}, \instructionround_{i,2}, \ldots, \instructionround_{i,\ell_i})$, where $\ell_i$ denotes the number of instructions and responses. 
The tool call sequence $\tool_i$ consists of tool calls $\tool_i=\bra{\toolusage_{i,1}\bra{\varphi_{i,1}}, \toolusage_{i,2}\bra{\varphi_{i,2}}, \ldots, \toolusage_{i, q_i}\bra{\varphi_{i,q_i}}}$, where $\toolusage_{i,j}\in\functionset{}$ corresponds to an executable function, $\functionset{}$ denotes the set of all possible tools, $\varphi_{i,j}$ represents the input to $\toolusage_{i,j}$, and $q_i$ is the total number of tool calls in the trajectory. %
Note that instructions and responses, tool calls, and outputs can be interleaved in time; in a slight abuse of notation, we use $\sample_i$ to refer to the time-ordered sequence of events, and use $\instruction_i, \tool_i, \finaloutput_i$ to refer to the corresponding filtered subsequences containing only instructions and responses, tool calls, and final outputs, respectively.

\vspace{-1.75mm}
\paragraph{\textbf{Summary of inputs}} \name{} takes as input (1) a real dataset $\base$; (2) a synthetic dataset $\base'$; and (3) optional metric configurations that may specify data-specific prior knowledge and agents $A_1, \ldots, A_h$ (details below). %
Both datasets contain instructions and responses, reference tool calls and outputs; 
\name{} also supports datasets missing responses, tool calls, or outputs.

\subsection{Evaluation Metrics} 

\name{} divides evaluation metrics into three pillars: validity (\cref{sec:validity}), fidelity (\cref{sec:fidelity}), and diversity (\cref{sec:diversity}).
\revise{Each pillar includes multiple sub-metrics that require varying amounts of prior knowledge about the real data, and evaluate different aspects of the synthetic dataset: (1) task instructions and responses, (2) tool calls, (3) outputs, and (4) downstream tasks. } 

\subsubsection{Validity Metrics}
\label{sec:validity}
Validity is important to assess since synthetic data can appear faithful yet be unusable if the tool calls or final outputs fail to complete the task. In practice, invalid samples may arise from hallucinated tool names or arguments, or from plausible-looking outputs that do not satisfy the instructions.

While validity may be dataset-specific, we use a broad definition: whether the provided tool calls or outputs accomplish the task instruction, serving as a basic self-consistency check.
By default, we assess validity for each sample $D_i \in \mathcal D$ \revise{with an LLM-as-a-judge, whose prompt and agreement with human annotations are reported in \cref{sec:app_valid_prompt}; users may instead define rule-based checkers when available.
} 
We report the overall \textbf{Validity Rate (\successrate)} as the proportion of valid tool-call sequences and outputs.

\subsubsection{Fidelity Metrics}
\label{sec:fidelity}
We design fidelity metrics to measure similarity between  synthetic and real data, which is especially useful when synthetic data is used as a direct replacement for real data, e.g., due to privacy constraints. 
Fidelity is the only property in \name{} that uses real data to compute its associated metrics, covering (1) task instructions and responses, (2) tool calls, (3) outputs, and (4) downstream tasks.

\vspace{-1mm}
\paragraph{\textbf{1. Fidelity metrics for task instructions and responses}} Task instructions and responses have an alternating structure, with each instruction followed by a response. Due to this, we draw inspiration from StructBench \citep{wangstruct}, a framework for evaluating structured synthetic data, and use metrics below.

For 
fidelity metrics that capture the structural relations in the data, %
we use \textit{Key Node Dependency} and \textit{Attribute Match}. \textbf{Key Node Dependency (\KND{})} measures semantic dependencies between parts of each sample by computing embedding cosine similarities between each instruction and its corresponding response, and between each response and the subsequent instruction. We then compare real and synthetic datasets by measuring the distributional distance between their similarity-score distributions.
\textbf{Attribute Match (\AM{})} measures how closely real and synthetic datasets align on predefined statistical and semantic attributes by computing distributional distances over their attribute distributions, using Wasserstein-2 distance for numerical attributes and Total Variation (TV) distance for categorical ones. We consider attributes including the number of instruction turns, instruction and response token lengths, and dataset-specific semantic properties (specified in \cref{tbl:metrics}). 

For non-structural fidelity metrics, we use \textbf{\precision} and \textbf{\recall} to measure the semantic quality and coverage. \precision{} (respectively, \recall{}) is the fraction of synthetic (respectively, real) samples whose embedding distance to a real (respectively, synthetic) sample is smaller than the distance to the $k$-th nearest neighbor within their own distribution. We also use \textbf{Fréchet Inception Distance (\fid{})} to measure semantic closeness between real and synthetic data.

\vspace{-1mm}
\paragraph{\textbf{2. Fidelity metrics for tool calls}}
We evaluate whether synthetic tool-usage and planning patterns match the real dataset. Each metric computes a distributional distance between real and synthetic data for a specific property.
(1) \textbf{Tool Usage Match (\tooluse{})} compares overall tool-usage patterns, $\text{\tooluse{}}=\dis(\distribution_{\toolusage},\distribution'_{\toolusage})$, where $\distribution_{\toolusage}$ and $\distribution'_{\toolusage}$ denote the tool-usage distributions in the real and synthetic data, respectively, and $\dis$ is the TV distance.
(2) \textbf{Tool Call Number Match (\toolnumber{})} compares the distributions of tool-call counts $q_i$ per sample $D_i$, $\text{\toolnumber{}}=\dis(\distribution_q,\distribution'_q)$, where $\dis$ is the Wasserstein-2 distance and $\distribution_q$ is the distribution over the tool-call number.  
(3) \textbf{k-Step Tool Planning Match (\toolplanning)} measures tool-planning similarity by comparing conditional distributions of the next tool given the previous $k-1$ tool calls, i.e., $\distribution_{\toolusage \mid \toolusage_1\cdots \toolusage_{k-1}}, \forall {\toolusage_1\cdots \toolusage_{k-1}\in \functionset{}^{k-1}}$. It is a weighted sum of TV distances between real and synthetic conditional distributions: 
\begin{align*}
\text{\toolplanning}=\sum_{\toolusage_1\cdots \toolusage_{k-1}\in \functionset{}^{k-1}} n_{\toolusage_1\cdots \toolusage_{k-1}} \cdot \dis\bra{\distribution_{\toolusage | \toolusage_1\cdots \toolusage_{k-1}}, \distribution'_{\toolusage | \toolusage_1\cdots \toolusage_{k-1}}},
\normalsize
\end{align*}
where $n_{\toolusage_1\cdots \toolusage_{k-1}}$ is the number of $k$-step tool-call sequences in the real data with prefix $\toolusage_1\cdots \toolusage_{k-1}$. \revise{We set $k\in\brc{1, 2}$ by default and allow larger values for long-horizon tasks.}

\vspace{-1mm}
\paragraph{\textbf{3. Fidelity metrics for outputs}} We evaluate the task fidelity of the final outputs by measuring their \textbf{\precision{}}, \textbf{\recall{}}, and \textbf{\fid{}} {relative to the real dataset}.

\vspace{-1mm}
\paragraph{\textbf{4. Fidelity metrics for downstream evaluation}} To assess the downstream utility of a benchmark constructed from synthetic data, we compare agent performance on real and synthetic tasks, measuring whether the two benchmarks induce similar tool-call and output-generation behavior. 
Recall that the user may optionally provide one or more agents $A_1,\ldots,A_h$ for the downstream evaluation; these agents can differ from the one being evaluated in \cref{fig:scenario}. For each agent $A_j$ and reference trajectory $D_i$, the agent generates tool calls at each instruction turn, conditioned on the current instruction, prior instruction turns, and the history of executed tool calls and results, using the dataset-specific tool set. The final output is then generated from the full instruction and executed tool-call history.

We adopt two downstream metrics. \textbf{(1) Task Difficulty Difference (\difficulty{})} measures the average absolute difference in task-completion performance across agents between real and synthetic tasks. For each reference trace $D_i$ from either dataset and each input agent $A_j$, we generate an agent trace $\tilde D_{i,A_j}$ and use an LLM-as-a-judge to determine whether $D_i$ and $\tilde D_{i,A_j}$ are functionally equivalent (prompt in \cref{sec:app_llm_judge}). We then compute the absolute difference in task success rates for tool-call selection or final-output generation between the real and synthetic benchmarks, averaged across agents. \textbf{(2) Ranking Divergence (\ranking{})} measures whether agent performance rankings are preserved across the real and synthetic datasets by computing the Spearman rank correlation between rankings based on tool-call or output correctness. Lower \difficulty{} and higher \ranking{} indicate better downstream utility.

\subsubsection{Diversity Metrics}
\label{sec:diversity}

{Diversity is particularly important when synthetic data is used to improve real-dataset coverage. 
For example, the real dataset may be collected in a controlled setting with little representation of outliers  or real-world user requests; practitioners may thus add synthetic outliers to test product robustness.}

{Measuring diversity is challenging: without prior domain-specific knowledge, it is unclear whether a dataset adequately covers the (high-dimensional) space of data traces.}
We therefore use two reference-free metrics, Vendi Score and Attribute Diversity, which do not require access to real data, though users of \name{} can compare the diversity of real and synthetic data if desired.

\textbf{Vendi Score (Vendi)} \cite{friedman2022vendi} is widely used to measure diversity in machine learning, including data curation and generative modeling \cite{jiang2025t2i,wu2023self}. 
It computes diversity from a similarity matrix $\smatrix$: for instructions \& responses and final outputs, $\smatrix_{i,j}$ is the cosine similarity between embeddings of $\instruction_i,\instruction_j$ or $\finaloutput_i,\finaloutput_j$, respectively, using \texttt{text-embedding-3-small} in our experiments. 
For tool calls, we define
$
\smatrix_{i,j} = 1 - \frac{\text{Levenshtein}\bra{\tool_i, \tool_j}}{\max\bra{q_i, q_j}}, 
$
where $q_i$ is the number of tool calls in $\tool_i$, and we adopt Levenshtein distance to measure sequence dissimilarity. Vendi then computes the exponential entropy of the eigenvalues $\lambda_1, \ldots, \lambda_m$ of the normalized matrix $\smatrix / m$:
$
\text{Vendi} = \text{exp}\bra{-\sum_{i=1}^{m} \lambda_i \log \lambda_i}.
$
Higher Vendi Score indicates more dissimilar samples and thus greater diversity; it is upper-bounded by $m$, achieved when all pairwise similarities are zero.
\revise{Vendi Score is agnostic to dataset format and content but may be harder to interpret. For users with domain knowledge, we also include Attribute Diversity, which is more interpretable but requires specifying attributes of interest.}

\textbf{Attribute Diversity (\entropy)} measures instruction-response diversity using user-specified attributes that capture dimensions the operator wants to diversify, such as attraction type and city in T1 or anomalous vs. benign behavior in a computer security context. 
Each sample is assigned, by a human or LLM, to one attribute-value combination, {e.g., (``attraction type=museum", ``city=Austin")}. %
We then compute the entropy of the resulting attribute distribution:
$
\entropy = -\sum_{i=1}^C p_i \log p_i,
$
where $p_i$ is the proportion of samples in attribute combination $i$, and $C$ is the number of possible combinations.  %
Higher \entropy{} indicates that no single attribute value dominates the population, reflecting greater diversity; it is upper-bounded by $\log C$, attained when all combinations are equally represented.

\name{} metrics and examples of dataset-specific prior knowledge are summarized in \cref{tbl:metrics}.

\section{Experiments}
\label{sec:experiment}

We demonstrate \name{} on the following baseline datasets and synthetic data generation methods. %

\vspace{-1.5mm}
\paragraph{\textbf{Evaluation datasets}}
We evaluate \name{} on T1 \citep{chakraborty2025t1}, BFCL \citep{patilberkeley}, and ACP \citep{kokel2025acpbench} benchmark datasets, and treat them as the real data. 
\textbf{T1} uses the T1-attraction dataset with 225 samples, where each sample contains multi-turn attraction-recommendation instructions, reference tool calls, and final outputs. \textbf{BFCL} uses the BFCL-V3-Base-Multi-Turn dataset with 200 samples, covering domains such as file operations, mathematical calculation, and travel booking, with reference tool calls. \textbf{ACP} uses the ACPBench-Applicability\&Progression dataset with 260 samples, covering planning domains such as transportation and robot action planning. Metrics across datasets are summarized in \cref{tbl:metrics}, \revise{and the computational costs of \name{} across datasets are reported in \cref{app:cost}.} %

\begin{table}[htbp]
  \caption{Evaluation metrics across datasets. Metrics in \orange{orange} require dataset-specific prior knowledge; ``---'' indicates no input is used.  %
  }
  \label{tbl:metrics}
  \centering
\scalebox{0.63}{
\begin{tabular}{|c|c|c|c|c|c|}
\Xhline{1.2pt}
\textbf{Evaluation Aspect}
& \textbf{Evaluation Target}
& \textbf{Metric}
& \textbf{T1 Instantiation}
& \textbf{BFCL Instantiation}
& \textbf{ACP Instantiation} \\
\Xhline{1.2pt}

\multirow{2}{*}{\textbf{Validity}}
& Tool Call
& Validity Rate
& LLM-as-a-judge
& LLM-as-a-judge
& N/A \\
\cline{2-6}
& Output
& Validity Rate
& LLM-as-a-judge
& N/A
& {LLM-as-a-judge} \\
\Xhline{1.2pt}

\multirow{13}{*}{\textbf{Fidelity}}
& \multirow{5}{*}{Instruction}
& Key Node Dependency
& \makecell{
1. (instruction, response) pair\\
2. (response, instruction) pair
}
& \makecell{(instruction, instruction) pair}
& \makecell{(context, instruction) pair} \\
\cline{3-6}

&
& Attribute Match
& \makecell{
1. number of instructions\\
2. instruction token length\\
3. response token length\\
\orange{4. city}\\
\orange{5. attraction type}
}
& \makecell{
1. number of instructions\\
2. instruction token length\\
\orange{3. task domain}\\
\orange{4. task subdomain}
}
& instruction token length \\
\cline{3-6}

&
& \precision
& ---
& ---
& --- \\
\cline{3-6}
&
& \recall
& ---
& ---
& --- \\
\cline{3-6}
&
& FID
& ---
& ---
& --- \\
\cline{2-6}

& \multirow{3}{*}{Tool Call}
& Tool Usage Match
& ---
& ---
& N/A \\
\cline{3-6}
&
& Tool Call Number Match
& ---
& ---
& N/A \\
\cline{3-6}
&
& \toolplanning{}
& $k \in \brc{2,3}$
& $k \in \brc{2,3}$
& N/A \\
\cline{2-6}

& \multirow{3}{*}{Output}
& \precision
& ---
& N/A
& --- \\
\cline{3-6}
&
& \recall
& ---
& N/A
& --- \\
\cline{3-6}
&
& FID
& ---
& N/A
& --- \\
\cline{2-6}

& \multirow{2}{*}{Downstream Task}
& Task Difficulty Difference
& \makecell{1. Tool Call\\2. Output}
& Tool Call
& Output \\
\cline{3-6}
&
& Ranking Divergence
& \makecell{1. Tool Call\\2. Output}
& Tool Call
& Output \\
\Xhline{1.2pt}

\multirow{4}{*}{\textbf{Diversity}}
& \multirow{2}{*}{Instruction}
& Vendi Score
& ---
& ---
& --- \\
\cline{3-6}
&
& Attribute Diversity
& \orange{(city, attraction type) pair}
& \orange{(domain, subdomain) pair}
& \orange{planning domain} \\
\cline{2-6}
& Tool Call
& Vendi Score
& ---
& ---
& N/A \\
\cline{2-6}
& Output
& Vendi Score
& ---
& N/A
& --- \\
\Xhline{1.2pt}

\end{tabular}
}
\end{table}

\paragraph{\textbf{Synthetic data generation}}
\revise{To evaluate the \textbf{sensitivity} of \name{} to structured, controlled changes in synthetic data quality,}   we design interpretable synthetic data algorithms that modify original trajectories to model common problems in synthetic data (details below): %
\textit{Blank Filling}, \textit{Oversampling}, and \textit{In-Context Generation} models combined fidelity and diversity degradation, and \textit{Invalidation} models degraded validity. \revise{To evaluate \name{} on \textbf{industry-standard} synthetic data techniques, we use NVIDIA NeMo \cite{nvidia_nemo}, a synthetic data tool that is used in part to test agents.}

\textbf{(1) \textit{Blank Filling}:} We randomly mask tokens in the original instructions with probability $p$ and prompt a language model to fill them in (prompt in \cref{sec:app_syn_gen_prompts}). Larger $p$ masks more information, causing greater deviation from the original instructions. We set $p\in \brc{0.1, 0.3, 0.5, 0.7, 0.9}$.

\textbf{(2) \textit{Oversampling}:} We construct a synthetic dataset of fixed size $m$ by first selecting one instruction-response sequence $R \in \mathcal R$, where $\mathcal R \triangleq \cup_j R_j$ is the set of all such sequences in $\mathcal D$. We use $R$ for the first $rm$ samples and fill the remaining $(1-r)m$ samples by sampling without replacement from $\mathcal R \setminus \brc{R}$. Larger $r$ creates more duplicates and thus lower diversity. We set $r \in \brc{0.1, 0.3, 0.5, 0.7, 0.9}$.

\textbf{(3) \textit{In-Context Generation}:} We generate synthetic instructions and responses by prompting language models with $\ice{}$ in-context examples (prompt in \cref{sec:app_syn_gen_prompts}). When $\ice{}=0$, generation relies only on the prompt and is weakly grounded in the original data, reducing fidelity. When $\ice{}>0$, fixed in-context examples limit diversity, while randomly varying them improves the instruction-space coverage. We set $\ice{} \in \brc{0, 1, 3, 5}$ and consider both fixed and randomly sampled examples. 

\textbf{(4) \textit{Invalidation}:} We keep the original instructions but replace a fraction $\invalid$ of samples with modified tool calls or outputs, leaving the rest unchanged. For benchmark datasets with tool calls $\toolusage\bra{\varphi}$, namely T1 and BFCL, we alter each selected tool call by replacing its input $\varphi_{i,j}$ with an input $\varphi_{i,k}$ that does not satisfy the instruction. For datasets with final outputs $O$, namely T1 and ACP, we replace each selected output $O_i$ with an alternative output $O_k$ that does not satisfy the instruction. 

\textbf{(5) \textit{Industry-standard synthetic data}:}
\revise{We generate synthetic agent trajectories using NVIDIA NeMo Data Designer, based on the specified input schemas, constraints, and backend language model. We use \texttt{GPT-4o-mini}, \texttt{Nvidia-Nemotron-Nano-9B-v2}, \texttt{Mistral-24B-Instruct}, and \texttt{Llama3.1-8B-Instruct} as backends, and vary the temperature over $\brc{0.1, 0.3, 0.5, 0.7, 0.9}$.}

For methods that generate synthetic instructions and responses, namely Blank Filling, Oversampling, and In-Context Generation, we also need corresponding tool calls and outputs, which we generate using \texttt{Llama3.1-8B-Instruct} as the backend for each dataset-specific agent.

\paragraph{\textbf{Agents for downstream evaluation}}
We conduct downstream evaluations on each real or synthetic dataset using LLM agents. For each trace, an agent iterates over the instruction sequence, receiving the full prior context, including previous instructions, responses, tool calls, and results, and predicts the next response and/or tool call, which is then compared with the provided trace. We evaluate three agents using the same benchmark runtime context, with backends \texttt{gemma-3-1b-it}, \texttt{Qwen3-4B-Instruct}, and \texttt{Llama3.1-8B-Instruct}. We use \texttt{Mistral-7B-Instruct} as the LLM-as-a-judge for functional equivalence of tool calls and outputs (prompt in \cref{sec:app_llm_judge}).

\paragraph{\textbf{Metric visualization}}
We visualize synthetic-data performance across metrics using line and radar plots. For radar plots, all metrics are rescaled to $[20,100]$, where $20$ is the worst performance among compared methods and $100$ is the metric upper bound (e.g., \precision{} = 1 or \toolplanningtwo{} = 0). To summarize \textbf{overall diversity, fidelity, and validity}, we compute an aggregate score for each by first normalizing the corresponding metrics to $[0,1]$ (with higher values indicating better performance) and then averaging them.
Although averaging normalized fine-grained metrics is not necessarily the most meaningful way as they may scale differently, this practice is sometimes used in benchmarks for easier visualization and interpretation \cite{wang2023decodingtrust}.

\subsection{Experimental Results}

\revise{We present the evaluation results of synthetic data generation methods proposed in this work on T1 in \cref{fig:plot_bar,fig:radar_diversity,fig:radar,fig:invalid,fig:fidelity_diversity}, with detailed results on all datasets in \cref{sec:app_result} and NVIDIA NeMo results on T1 in \cref{sec:app_nemo}.}

\subsubsection{\name{} can capture fine-grained variations in fidelity, diversity, and validity of synthetic data.}

\begin{figure}[htbp]
    \centering
\begin{subfigure}{0.43\textwidth}
         \centering
    \includegraphics[width=1\linewidth]{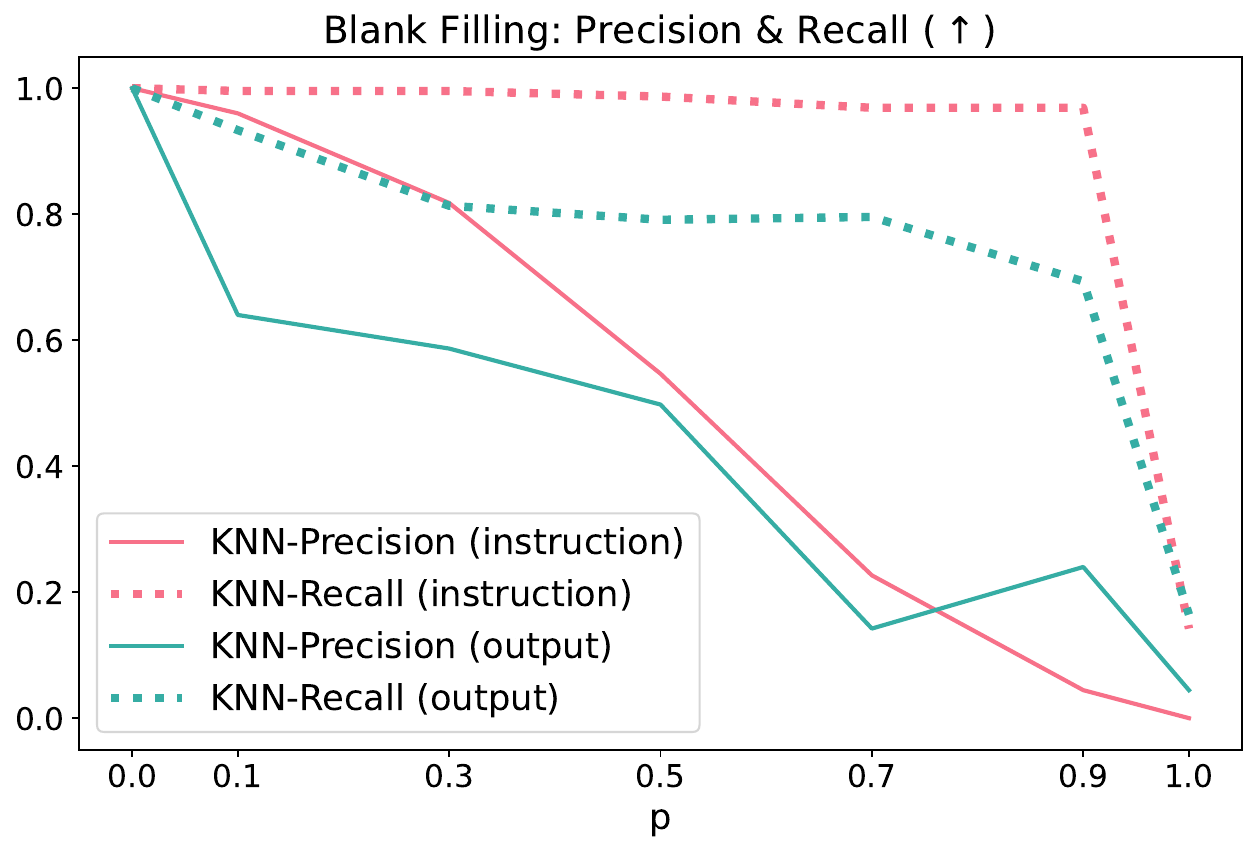}
    \caption{\precision{} and \recall{} of Blank Filling with different masking probability $\blankfilling$.}
    \label{fig:blankfilling_precision}
\end{subfigure}
\quad
\begin{subfigure}{0.47\textwidth}
         \centering
    \includegraphics[width=1\linewidth]{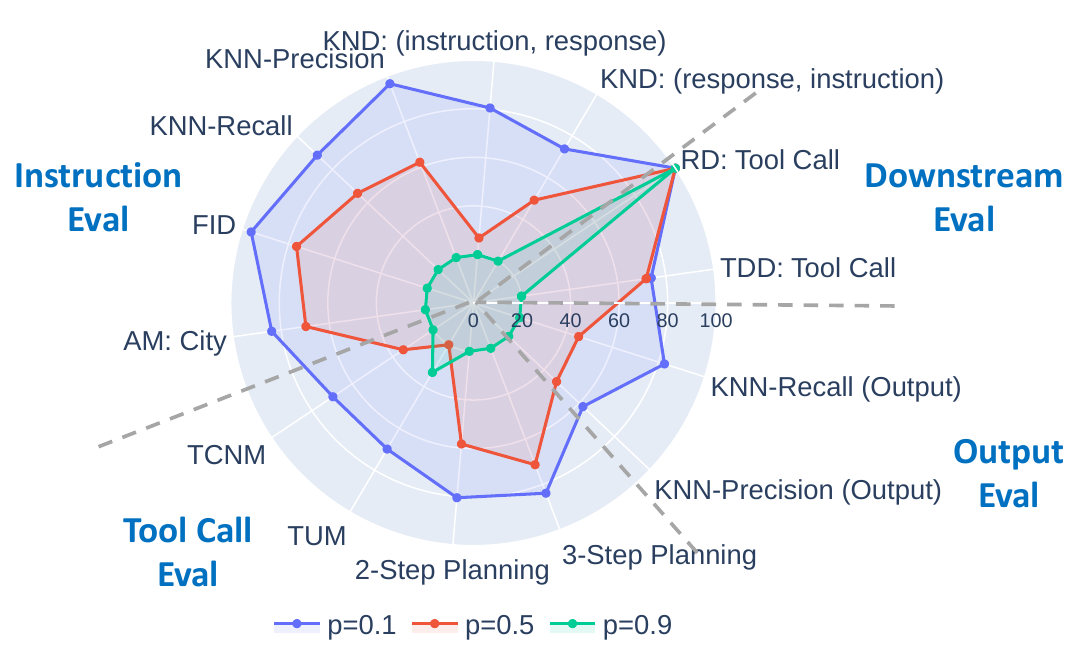}
    \caption{Blank Filling. As $\blankfilling{}$ increases, data fidelity degrades across nearly all metrics.}
    \label{fig:blank_filling_radar}
\end{subfigure}
\begin{subfigure}{0.43\textwidth}
         \centering
    \includegraphics[width=1\linewidth]{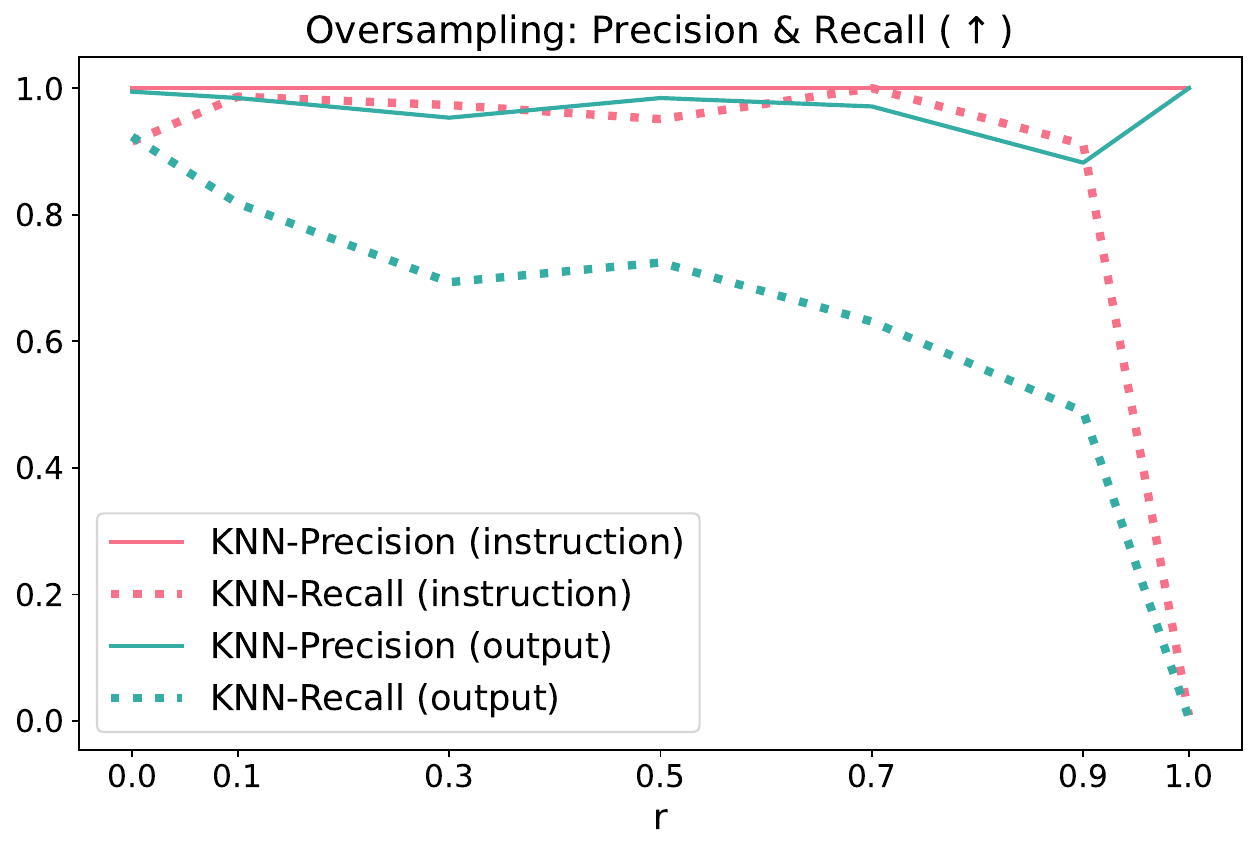}
    \caption{\precision{} and \recall{} of Oversampling with different duplication rate $\oversampling$.}
    \label{fig:oversampling_precision}
\end{subfigure}
\quad
\begin{subfigure}{0.47\textwidth}
         \centering
    \includegraphics[width=1\linewidth]{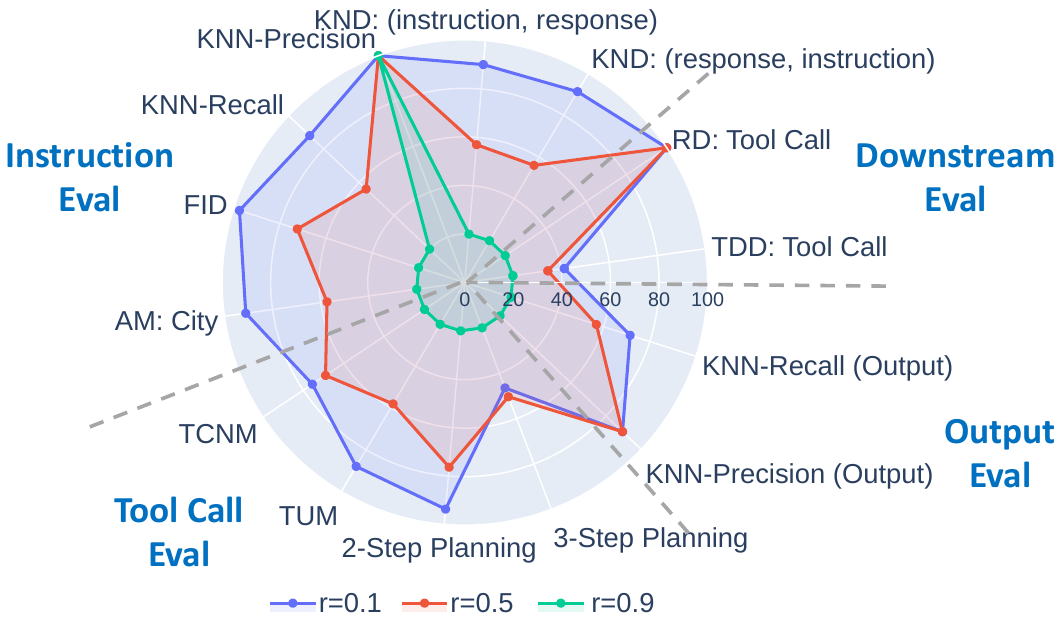}
    \caption{Oversampling. As $\oversampling{}$ increases, data fidelity degrades across nearly all metrics. %
    }
    \label{fig:oversampling_radar}
\end{subfigure}
\caption{Fidelity of Blank Filling and Oversampling on the T1 dataset. %
}
\label{fig:plot_bar}
\vspace{-1mm}
\end{figure}

\noindent \textbf{Fidelity:} \cref{fig:plot_bar} shows the fidelity of Blank Filling and Oversampling on T1.
For Blank Filling (\cref{fig:blankfilling_precision}), as $\blankfilling{}$ increases, the \precision{} for both instructions and outputs decreases from nearly $1$ to close to $0$, indicating progressive degradation in semantic quality, consistent with the intuition that higher masking probability lowers fidelity. In contrast, when $\blankfilling{}\leq 0.9$, the \recall{} for both instructions and outputs remains relatively high (above $0.7$), suggesting that Blank Filling largely preserves semantic coverage. 
For Oversampling (\cref{fig:oversampling_precision}), increasing $\oversampling{}$ significantly reduces output \recall{}, while instruction and output \precision{} remain near perfect, consistent with the intuition that duplication reduces task coverage while largely preserving task quality. 
Moreover, \cref{fig:blank_filling_radar,fig:oversampling_radar} show that increasing $\blankfilling{}$ or $\oversampling{}$ degrades performance across nearly all fidelity metrics, aligning with the intuition that heavier masking or duplication produces lower-fidelity synthetic data.

\begin{figure}[htbp]
    \centering
\begin{subfigure}{0.48\textwidth}
         \centering
    \includegraphics[width=1\linewidth]{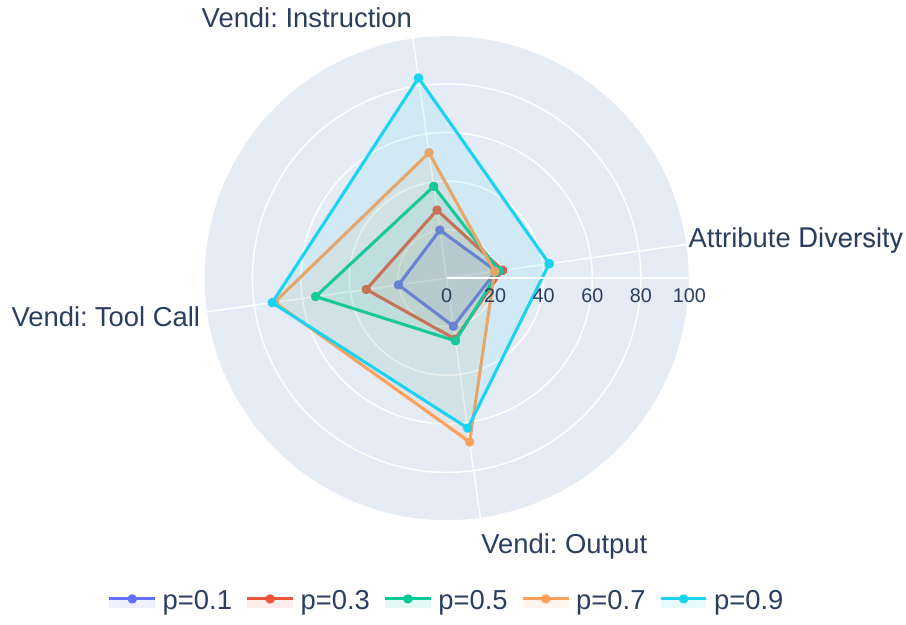}
    \caption{Vendi Scores and Attribute Diversity of Blank Filling with different masking probability $\blankfilling$. As $\blankfilling$ increases, data diversity improves.}
    \label{fig:blankfilling_diversity}
\end{subfigure}
\quad
\begin{subfigure}{0.48\textwidth}
         \centering
    \includegraphics[width=1\linewidth]{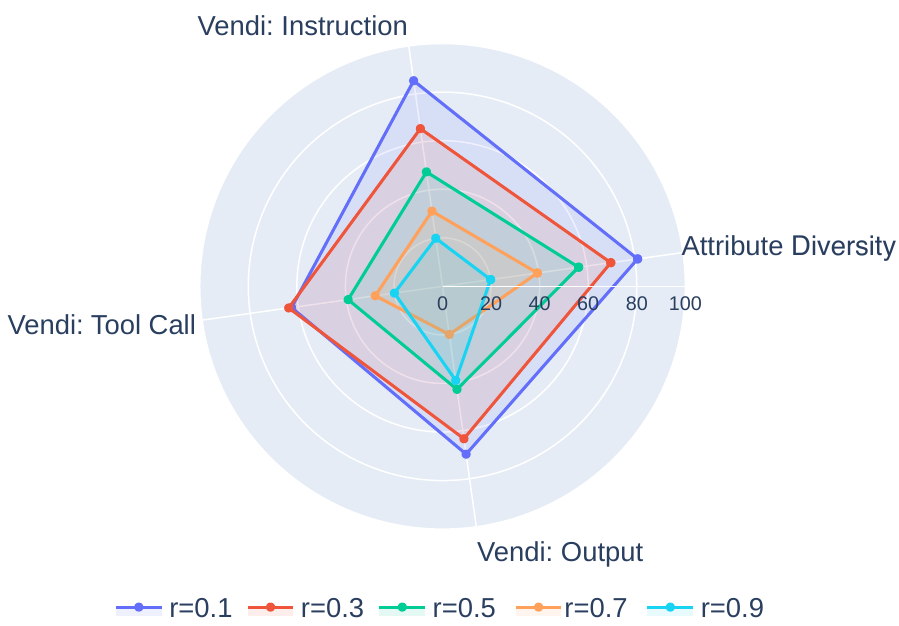}
    \caption{Vendi Scores and Attribute Diversity of Oversampling with different duplication rate $\oversampling$. As $\oversampling{}$ increases, data diversity decreases.}
    \label{fig:oversampling_diversity}
\end{subfigure}
\caption{Diversity metrics for Blank Filling and Oversampling on the T1 dataset. %
}
\label{fig:radar_diversity}
\end{figure}

\noindent \textbf{Diversity:}
\cref{fig:radar_diversity} shows the diversity of Blank Filling and Oversampling on T1. As the masking probability $\blankfilling$ increases, Blank Filling becomes more diverse (see \cref{fig:blankfilling_diversity}), consistent with the intuition that masking more words introduces novel content and diversifies task topics and execution trajectories. In contrast, Oversampling diversity decreases as the duplication rate $\oversampling$ increases (see \cref{fig:oversampling_diversity}), consistent with the intuition that more duplicated samples reduce overall dataset diversity.

\begin{wrapfigure}[15]{r}{0.45\textwidth}
    \centering
    \includegraphics[width=\linewidth]{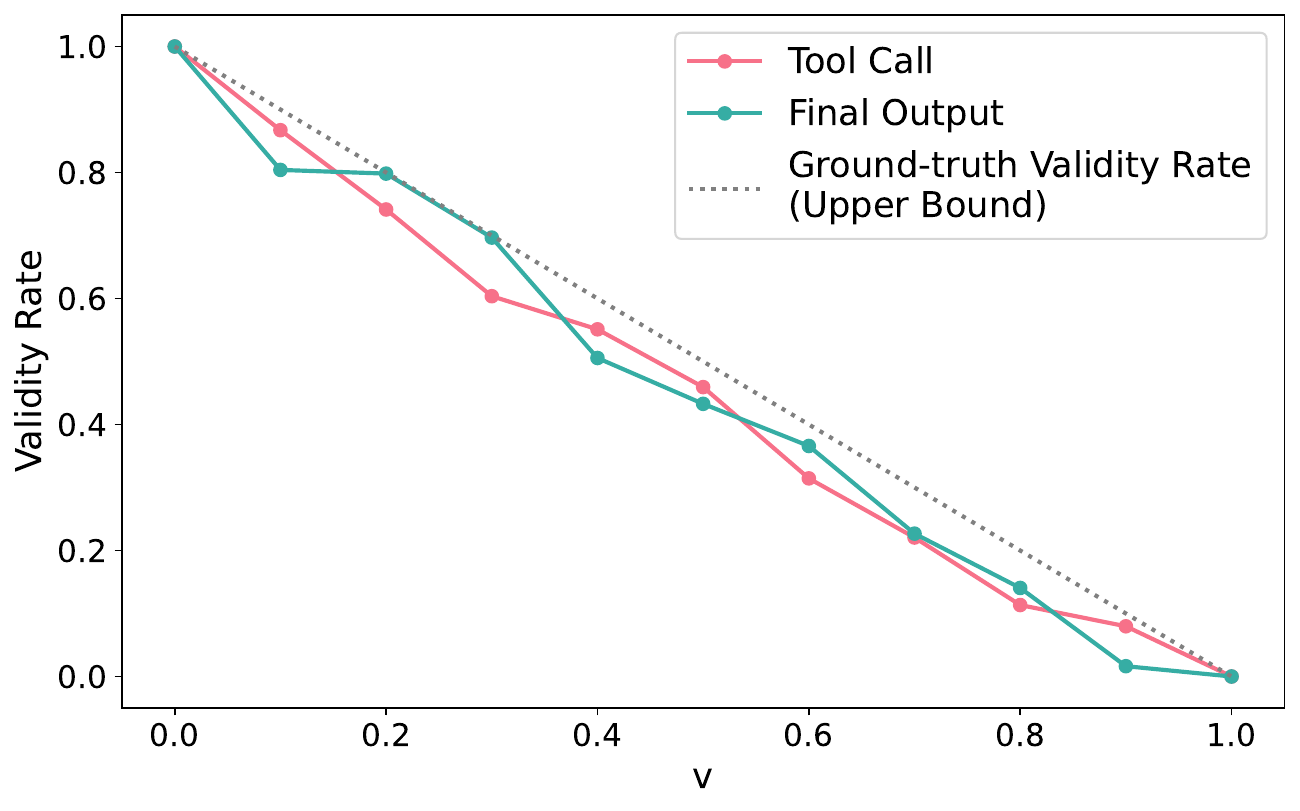}
    \caption{Validity of Invalidation on T1. As invalidation ratio $\invalid$ increases, Validity Rates for both tool calls and outputs decrease.
    }
    \label{fig:invalid}
\end{wrapfigure}

\noindent \textbf{Validity:}
\cref{fig:invalid} illustrates the Validity Rate %
of \emph{Invalidation} on T1. As the invalidation ratio $\invalid$ increases, Validity Rates for both tool calls and final outputs decrease monotonically with an approximate slope of $-1$, as expected by construction. %
This indicates that \name{} effectively captures validity variations. %

\subsubsection{No single metric can fully characterize synthetic data performance.} 
\begin{wrapfigure}{l}{0.5\textwidth}
    \centering
    \includegraphics[width=\linewidth]{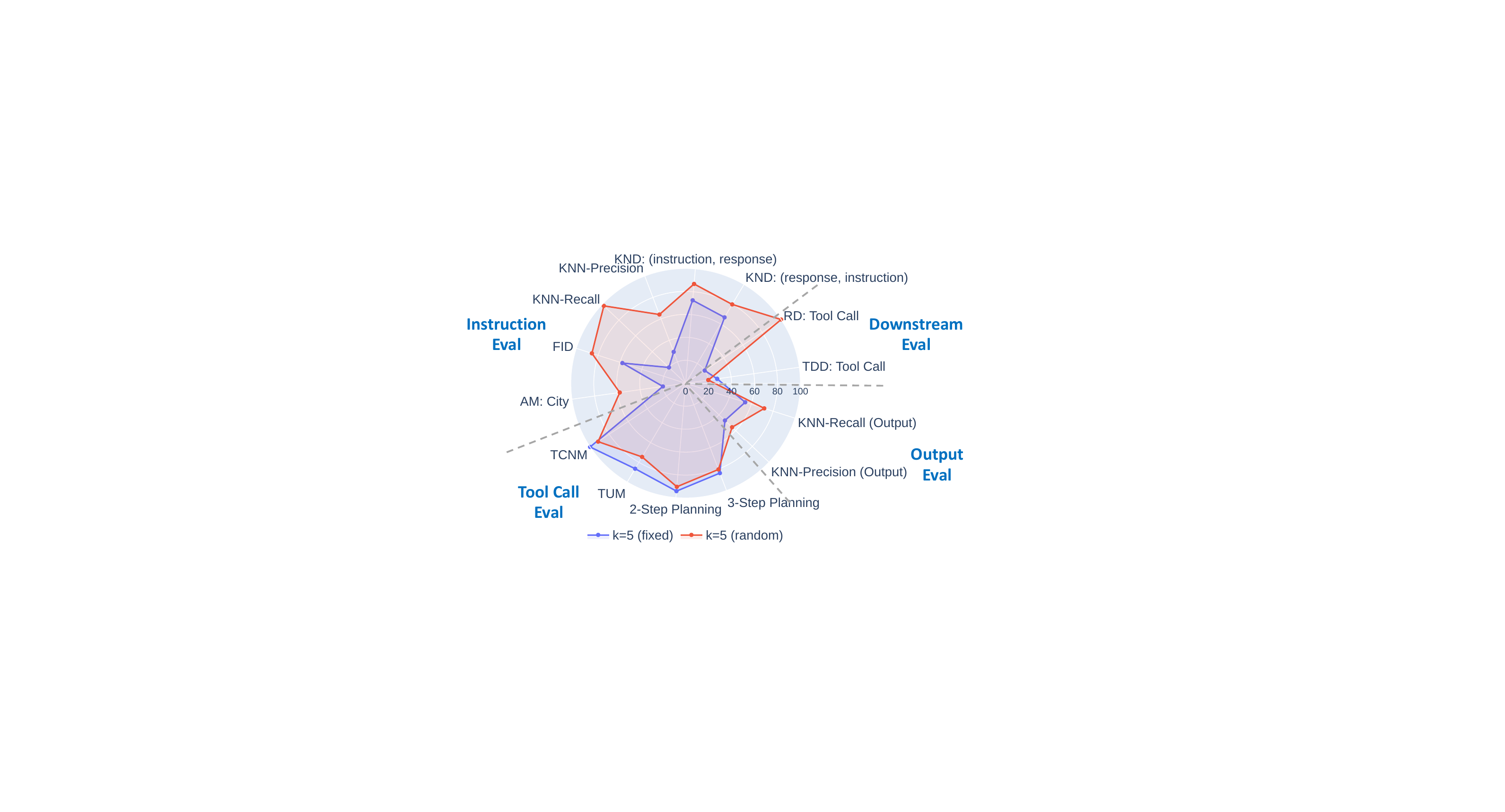}
    \caption{Fidelity vs. diversity for Blank Filling and Oversampling. For Blank Filling, a higher masking probability $\blankfilling$ leads to lower fidelity but greater diversity. For Oversampling, a higher duplication rate $\oversampling$ results in declines in both fidelity and diversity.
    }
    \label{fig:fidelity_diversity}
\end{wrapfigure}

\cref{fig:fidelity_diversity} illustrates the fidelity-diversity trade-offs for Blank Filling and Oversampling. For each dimension, we compute an aggregate score to summarize overall performance. %
Increasing the duplication rate $\oversampling$ in Oversampling degrades both fidelity and diversity, whereas increasing the masking probability $\blankfilling$ in Blank Filling lowers fidelity but improves diversity. These results highlight that no single metric category fully characterizes synthetic data performance, as different metrics capture distinct aspects.

\begin{figure}[htbp]
    \centering
    \vspace{0mm}
\begin{subfigure}{0.45\textwidth}
         \centering
    \includegraphics[width=1\linewidth]{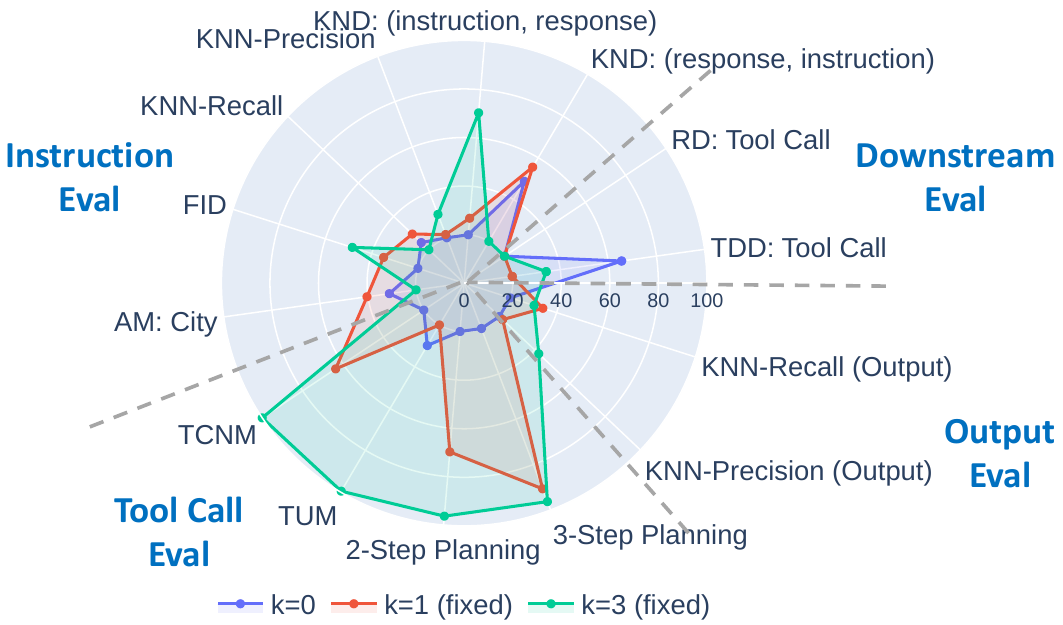}
    \caption{In-Context Generation with different number of in-context examples $\ice{}$. Increasing $\ice{}$ does not consistently improve all fidelity metrics, including \recall{}, \KND{}, and \AM{}.}
    \label{fig:ice_fix_radar}
\end{subfigure}
\quad
\begin{subfigure}{0.45\textwidth}
         \centering
    \includegraphics[width=1\linewidth]{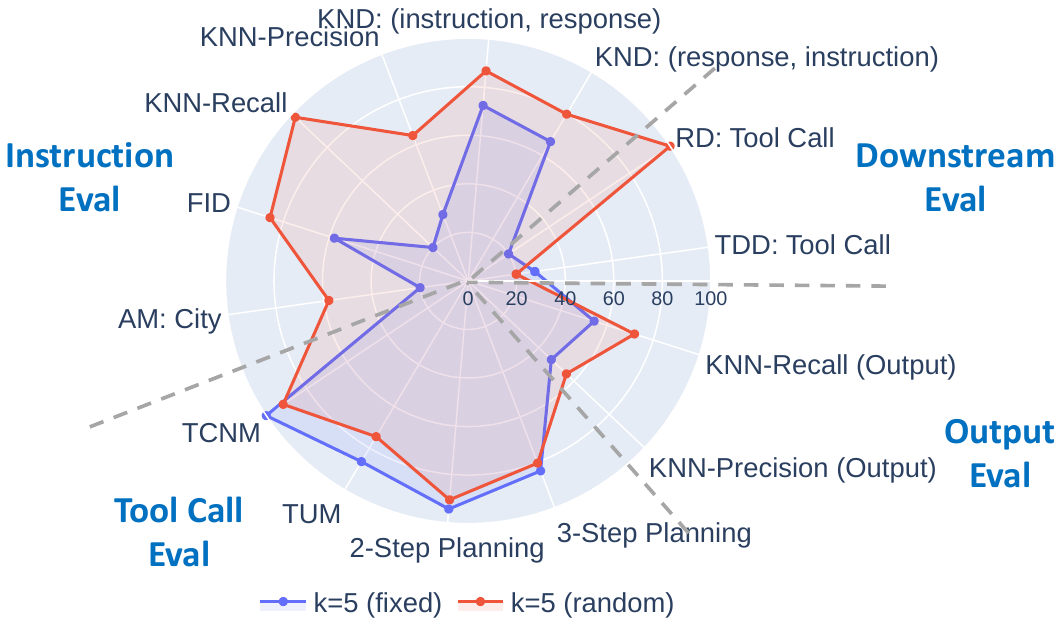}
    \caption{In-Context Generation with $\ice{}=5$. Randomized in-context examples improve instruction and output fidelity over fixed examples, while maintaining similar tool-call performance.}
    \label{fig:ice_random_radar}
\end{subfigure}
\caption{Fidelity of In-Context Generation under T1 with fixed or randomized in-context examples. %
}
\label{fig:radar}
\vspace{-2mm}
\end{figure}

Even within fidelity evaluation itself, one parameter setting may outperform another on some metrics while underperforming on others. For example, in In-Context Generation (\cref{fig:ice_fix_radar}), increasing the number of in-context examples $\ice{}$ does not uniformly improve all fidelity metrics: on T1, $\ice{}=3$ improves \precision{} over $\ice{}=1$ but lowers \recall{}.
Fixing $\ice{}$ and comparing fixed versus randomized examples (\cref{fig:ice_random_radar}), randomization improves instruction- and output-level fidelity while maintaining similar tool-call performance.
\revise{We also show in \cref{app:baseline} that simple baseline metrics, such as corpus-level statistics and embedding closeness, can be misleading when used alone.}
Another example appears under Blank Filling (\cref{fig:blank_filling_radar}): as $\blankfilling{}$ increases, tool-call metrics degrade, indicating larger discrepancies in tool-planning patterns between real and synthetic tasks. In contrast, the downstream metric \ranking{}: Tool Call remains strong, %
since higher-capacity models continue to outperform lower-capacity ones even when planning patterns shift, resulting in largely consistent model rankings. These observations further motivate the need for a comprehensive evaluation framework such as \name{}.

\subsubsection{Case study: Practitioners can leverage \name{} to iteratively diagnose and improve synthetic data generation.}
\label{sec:esdae_case_study}
We present a case study simulating a common workflow: a data holder starts with skewed data, uses \name{} to diagnose issues, then iteratively refines an augmentation strategy with \name{}'s feedback.

We start with a modified version of T1 where some properties are under-represented: \revise{we select half of the attraction types, including culture, sport, culinarian, and guide, %
and downsample instances containing these types to $10\%$ of the original count by discarding the remaining samples.} The data holder then aims to augment the dataset to better represent these target types.

\noindent \textbf{Step 1: Diagnose the bottleneck with \name{}.}
The data holder first evaluates the skewed data with \name{}, which identifies \textit{diversity} as the primary issue (first column of \cref{tabl:case_study}) and pinpoints under-represented attributes via attribute distributions \revise{(reported in \cref{app:case_study})}, motivating targeted augmentation.

\noindent \textbf{Step 2: Attempt a lightweight fix via relabeling, then re-evaluate.}
As a fast intervention, the data holder applies \textit{Relabeling}, which replaces attraction-type keywords in the original instructions with target types. For example, we might change ``Please find some \textit{art} attractions in Canada'' to ``Please find some \textit{sporting} attractions in Canada.'' %
\name{} confirms that diversity improves (2nd column of \cref{tabl:case_study}); however, validity drops, revealing that naive keyword substitution can introduce semantic inconsistencies. For example, the modified sporting-attraction instruction may still be followed by an art-related response such as ``Do you prefer \textit{museums} or \textit{concerts}?''. This can make the modified instructions incompatible with the tool calls and outputs, causing failures in task completion.

\revise{\noindent \textbf{Step 3: Escalate to model-based synthesis via NVIDIA NeMo, then re-evaluate.}
To improve diversity while preserving validity, the data holder next adopts \textit{NVIDIA NeMo}. They first use smaller backend models, \texttt{Llama3.1-8B-Instruct} and \texttt{Nvidia-Nemotron-Nano-9B-v2}, with temperature 0.6, and provide prompts specifying the target attraction types. For example, given the target type \textit{sport}, NeMo generates the instruction ``Please find some \textit{sporting} attractions in Canada.'' followed by responses such as ``Do you prefer \textit{basketball} or \textit{baseball}?'', preserving semantic consistency. 
\name{} shows that NVIDIA NeMo with \texttt{Llama} or \texttt{Nemotron} preserves validity, but substantially sacrifices fidelity, while diversity remains moderate (columns 3 \& 4 of \cref{tabl:case_study}). %
The data holder then escalates to the higher-capacity backend \texttt{GPT-4o-mini}. \name{} shows that this improves
diversity while preserving validity, without substantially sacrificing fidelity (column 5 of \cref{tabl:case_study}).}

\begin{table}[htbp]
\centering
\caption{Validity, fidelity, and diversity of the skewed real data, Relabeling, and NVIDIA NeMo. 
\green{Green} indicates strong performance, \red{red} indicates poor performance, and black indicates moderate performance. The diversity of the original (unskewed) data is $0.72$.
}
\begin{tabular}{cccccc}
\toprule
& Skewed Real Data & Relabeling & \multicolumn{3}{c}{NVIDIA NeMo} \\
\cmidrule(lr){4-6}
& & & \texttt{Llama} & \texttt{Nemotron} & \texttt{GPT} \\
\midrule
Validity  & \green{1.0} & \red{0.82} & \green{0.98} & \green{0.99} & \green{0.99} \\
Fidelity  & \green{1.0} & \green{0.95} & \red{0.71} & \red{0.79} & 0.94 \\
Diversity & \red{0.48} & 0.65 & 0.67 & 0.61 & \green{0.70} \\
\bottomrule
\end{tabular}
\label{tabl:case_study}
\end{table}

\cref{tabl:case_study} summarizes fidelity, diversity, and validity using aggregate scores. 
We define ``strong" performance as metrics within 5\% of the original (unskewed) real dataset and ``poor" performance as metrics at least 15\% lower. %
Overall, this case study shows how \name{} supports iterative refinement: diagnose the limiting factor, apply an intervention, and use metric feedback to adjust the strategies.

\section{Conclusions and Limitations}
\label{conclusion}

In this paper, we introduced \name{}, a multi-axis framework for evaluating how well  synthetic data replicates and augments real agent trajectories across task instructions and responses, tool calls, final outputs, and downstream performance. Through experiments on recent agent benchmarks with realistic and controlled synthetic generation schemes, \name{} consistently detects fine-grained failures in validity, fidelity, and diversity. Overall, no single metric suffices to capture synthetic benchmark quality, motivating multi-axis evaluation with \name{} before using synthetic data for agent testing.
\revise{A limitation of our work is that \name{} mainly focuses on multi-turn tool-calling agentic settings. Extending the framework to agent benchmarks involving interactive environments or multi-agent coordination is an important direction for future work.}

\section*{Acknowledgments}
This work was supported in part by the NSF RINGS program, grant CNS-2148359.

\bibliographystyle{plain}

\newpage
\appendix

\section{Related Work}
\label{app:related_work}

Robust benchmarks for interactive tool-use  are necessary both for generalist agents (e.g., code generation, research assistance, and open-domain conversation \citep{yehudai2025survey, anthropic2026demystifying,yang2024swe,shao2024assisting,yao2022react,ouyang2022training}) and domain-specific agents with narrowly scoped workflows (e.g., in finance, technology, and corporate services \citep{pan2025measuring,enkryptai,mcgrath}).
However, constructing and maintaining such benchmarks is costly and requirements evolve \citep{mohammadi2025evaluation, yehudai2025survey, hutchinson2022evaluation}, motivating continuously updated and synthetic benchmark generation approaches \citep{white2025livebench, zhu2023dyval}.

\paragraph{\textbf{Evaluating synthetic benchmarks for LLMs.}} 
There have been several synthetic benchmark datasets for evaluating sequential,  conversational interactions between a user and an LLM; a few papers have quantitatively evaluated the quality of these benchmarks \cite{gill2025has,maheshwari2024efficacy,xiong2025probe,majurski2025grounding}. 
However, these lines of work do not tackle \emph{agent} benchmarks, %
where evaluation is more complicated due to multi-step decision making, interaction with environments/tools, and the need to attribute failures to specific components of an agentic pipeline.
Examples include \cite{gill2025has,maheshwari2024efficacy,majurski2025grounding}, which study synthetic benchmarks for standard NLP tasks (e.g., reading comprehension, intent detection). They assess synthetic data via factors like difficulty (i.e., whether an agent has a similar completion rate on tasks in the real/synthetic data) and whether model rankings are preserved. 
\name{} generalizes these ideas to the multi-turn tool-calling setting in the fidelity evaluation of downstream tasks.
Relatedly, \cite{xiong2025probe} and \cite{mehta2025prompt} demonstrate that LLM behavior---both response content and trustworthiness---can differ between synthetic-benchmark settings and real-world deployment, motivating the need for realistic synthetic data. 

\paragraph{\textbf{Evaluating synthetic benchmarks for tool-calling agents.}} 
To our knowledge, prior works on evaluating synthetic benchmarks for tool-calling agents have only considered \emph{single-turn} benchmarks, in which each dataset instance consists of a single client query followed by a single agent response \citep{shen2024taskbench,iskander2024quality}.
This significantly simplifies evaluation, as it does not require the synthetic data to capture sequential dependencies among task instructions %
present in the real data; measuring the quality of these transitions is a major component of \name{}. 
Moreover, these prior works only evaluate the instructions (human inputs) in the synthetic data (e.g., naturalness, coherence), avoiding systematically evaluating the reference tool calls and outputs correspond to those instructions \citep{shen2024taskbench,iskander2024quality}. 
\cite{iskander2024quality} further evaluates synthetic samples by how much they help in-context learning, which serves as an indirect signal of data quality. Overall, these approaches do not provide quantitative metrics %
for systematically measuring the statistical and semantic properties of synthetic tasks directly, nor do they evaluate the associated tool calls and final outputs within task-completion trajectories.

Recent work has also proposed qualitative checklists for practitioners to evaluate benchmarks. For example, \cite{zhu2025establishing} advocate for verifying task validity, i.e., each sample in the benchmark should have the correct ground truth, while \cite{alonso2025evaluating} emphasize that a benchmark should contain realistic samples that test a diverse range of agent behaviors.
However, these checklists typically requires practitioners to manually review samples, making benchmark creation and validation time-consuming and iterative \citep{pan2025measuring}, especially when practitioners replace or augment real datasets with  synthetic samples \citep{white2025livebench,zhu2023dyval}.
\name{} speeds up these workflows by providing fine-grained quantitative metrics that pinpoint which parts of the benchmark can be improved. We demonstrate how \name{} can help practitioners through a case study in \cref{sec:esdae_case_study}.

\paragraph{\textbf{%
Evaluating multi-turn tool-calling agents}}
Multi-turn tool-calling agents are typically evaluated using end-to-end task success on interactive benchmarks, where agents execute multi-step tool calls and are scored by state-based checkers that compare the final environment or database state to an annotated goal state \citep{yao2024tau,liu2023agentbench,zhou2023webarena,yang2023intercode,xie2024osworld,jimenez2023swe}. While these works primarily evaluate \textit{agents} on a given benchmark, in contrast, we evaluate the \textit{benchmarks themselves}. 
{We assume text-based outputs in this work, but the \name{} framework could easily be extended to incorporate a state check as part of the output evaluation.}
Many benchmarks additionally enforce turn-level validity by jointly checking tool outputs and dialogue via state-based and response-based criteria across all turns \citep{patilberkeley,li2025toolrm,zhanggecko}.
Increasingly, evaluations also incorporate LLM-as-a-judge, using a strong judge model to determine pass/fail or preference-style win rates over full tool-use trajectories when multiple solutions are plausible \citep{qin2023toolllm,pan2024autonomous,xue2025illusion,lu2025agentrewardbench}.
\name{} also employs an LLM-as-a-judge to assess tool calls and final outputs by jointly checking them against the instructions across turns in our downstream evaluation. However, this alone is insufficient for evaluating the quality of a synthetic benchmark, as it focuses only on end-to-end agent performance and does not capture the intrinsic properties of task instructions or their associated tool calls and outputs. 
Hence, we introduce additional metrics that quantitatively measure the semantic and statistical properties of task instructions, tool-usage and planning patterns, and the quality and diversity of final outputs, providing richer diagnostic signals for developers. We demonstrate the value of these metrics in \cref{sec:experiment}.

\section{Prompts for LLM-as-a-Judge on Validity}
\label{sec:app_valid_prompt}

\subsection{Prompts for T1}

\subsubsection{Tool Call Evaluation}

System Prompt: 

\begin{lstlisting}
You are an evaluator. You must answer with ONLY 'yes' or 'no'. Never provide explanations or reasoning.
\end{lstlisting}

User Prompt: 

\begin{lstlisting}
Check the validity of the tool call sequence for an attractions recommendation assistant:

Conversation:
{instr}

Tool call:
{tool_call}

Evaluation rules:
- Tool call must accomplish the goal of the conversation
- Partial correctness = NO

Does the tool call correctly implement the conversation?
Answer (yes/no):
\end{lstlisting}

\subsubsection{Output Evaluation}

System Prompt: 

\begin{lstlisting}
You are an evaluator. You must answer with ONLY 'yes' or 'no'. Never provide explanations or reasoning.
\end{lstlisting}

User Prompt: 

\begin{lstlisting}
Check the validity of the output for an attractions recommendation assistant:

Conversation:
{instr}

Output:
{expected_output}

Evaluation rules:
- Output must accomplish the goal of the conversation
- Partial correctness = NO

Does the output correctly implement the conversation?
Answer (yes/no):
\end{lstlisting}

\subsection{Prompts for BFCL}

\subsubsection{Tool Call Evaluation}

System Prompt: 

\begin{lstlisting}
You are an evaluator. You must answer with ONLY 'yes' or 'no'.
Never provide explanations or reasoning.
\end{lstlisting}

User Prompt: 

\begin{lstlisting}
Check the validity of the tool call sequence for a function calling assistant:

User Requests:
{instr}

Tool calls:
{expected_tool_call}

Evaluation rules:
- Tool call must accomplish the goal of the conversation
- Partial correctness = NO

Does the tool call correctly implement the conversation?
Answer (yes/no):
\end{lstlisting}

\subsection{Prompts for ACP}

\subsubsection{Output Evaluation}

System Prompt: 

\begin{lstlisting}
You are an evaluator. You must answer with ONLY 'yes' or 'no'. Never provide explanations or reasoning.
\end{lstlisting}

User Prompt: 

\begin{lstlisting}
Check the validity of the output:

Conversation:
{instr}

Output:
{expected_output}

Evaluation rules:
- Output must accomplish the goal of the conversation
- Partial correctness = NO

Does the output correctly implement the conversation?
Answer (yes/no):
\end{lstlisting}

\subsection{\revise{Agreement Rate with Human Annotation}}

We conduct human annotation on 100 randomly selected synthetic T1 samples to assess whether the generated tool calls and outputs correctly fulfill the given tasks. We then compare the human labels with judgments produced by an LLM-as-a-judge using \texttt{Mistral-7B-Instruct}. The LLM judge achieves an F1 score of 0.86, with Cohen's $\kappa$ as 0.61, indicating substantial agreement with human annotations. This suggests that the LLM-as-a-judge provides a reliable default validity checker when task-specific rule-based checkers are unavailable.
\section{Prompts for LLM-as-a-Judge in Downstream Evaluation}
\label{sec:app_llm_judge}

\subsection{Prompts for T1}

\subsubsection{Tool Call Evaluation}

System Prompt: 

\begin{lstlisting}
You are an evaluator. You must answer with ONLY 'yes' or 'no'. Never provide explanations or reasoning.
\end{lstlisting}

User Prompt: 

\begin{lstlisting}
Compare these tool call sequences for an attractions recommendation assistant:

Conversation:
{instr}

Expected tool calls:
{expected_tool_call}

Actual tool calls:
{actual_tool_call}

Evaluation rules:
- Actual must accomplish the same goal as expected
- Semantic equivalence is OK (e.g., "OR" vs "Oregon", reordered operations with same result)
- Partial correctness = NO
- Different variable/cache names = OK if functionality identical

Does actual correctly implement the conversation based on expected?
Answer (yes/no):
\end{lstlisting}

\subsubsection{Output Evaluation}

System Prompt: 

\begin{lstlisting}
You are an evaluator. You must answer with ONLY 'yes' or 'no'. Never provide explanations or reasoning.
\end{lstlisting}

User Prompt: 

\begin{lstlisting}
Compare these outputs for an attractions recommendation assistant:

Conversation:
{instr}

Expected output:
{expected_output}

Actual output:
{actual_output}

Evaluation rules:
- Actual must convey same information and meaning as expected
- Different wording is OK if content equivalent
- Partial correctness = NO
- Focus on semantic content, not syntax

Does actual correctly respond to the conversation based on expected?
Answer (yes/no):
\end{lstlisting}

\subsection{Prompts for BFCL}

\subsubsection{Tool Call Evaluation}

System Prompt: 

\begin{lstlisting}
You are an evaluator. You must answer with ONLY 'yes' or 'no'.
Never provide explanations or reasoning.
\end{lstlisting}

User Prompt: 

\begin{lstlisting}
Compare these outputs for a function calling assistant:

User Requests:
{instr}

Expected tool calls:
{expected_tool_call}

Actual tool calls:
{actual_tool_call}

Evaluation rules:
- Actual must accomplish the same goal as expected
- Semantic equivalence is OK (e.g., reordered operations with same result)
- Partial correctness = NO
- Different variable names = OK if functionality identical

Does actual correctly implement the user requests based on expected?
Answer (yes/no):
\end{lstlisting}
\section{\revise{Computational Cost of \name{} Across Datasets}}
\label{app:cost}

For the LLM-as-a-judge in both validity checks and downstream evaluations, we use \texttt{Mistral-7B-Instruct}. For the agents in downstream evaluations, we use \texttt{gemma-3-1b-it}, \texttt{Qwen3-4B-Instruct}, and \texttt{Llama3.1-8B-Instruct} across all datasets.

\begin{itemize}
\item T1 (225 samples): 450 LLM calls for validity checking of tool calls and outputs; 1,350 LLM calls for downstream evaluation per agent.

\item BFCL (200 samples): 200 LLM calls for validity checking of tool calls; 800 LLM calls for downstream evaluation per agent.

\item ACP (260 samples): 260 LLM calls for validity checking; 0 LLM calls for downstream evaluation per agent, since agent responses can directly be compared with benchmark ground truth (answers are either `yes' or `no').
\end{itemize}

All experiments use open-source LLMs. For reference, using GPT-5.4-mini instead would cost under \$5 per dataset.

\section{Prompts for LLM-Based Synthetic Data Generation Methods}
\label{sec:app_syn_gen_prompts}

\subsection{Prompts for T1}

\subsubsection{Blank Filling}
\label{sec:app_t1_blankfill_prompts}

\noindent System Prompt: 

\begin{lstlisting}
You are a helpful conversation generator. When given a conversation with blanks (underscores), fill them in naturally. IMPORTANT RULES:
1. The conversation MUST start with 'assistant:' (not 'Assistant:' or any variation)
2. Lines MUST alternate strictly between 'user:' and 'assistant:'
3. Each line must follow the format: 'role: content' where role is either 'user' or 'assistant'
4. Output ONLY the completed conversation with no preamble, explanation, or extra text
5. Maintain the same number of conversation turns as the input
\end{lstlisting}

\noindent User Prompt: 

\begin{lstlisting}
Example input for fill in the blanks:

assistant: H_____ What ____ of attractions are you looking for? Are you interested in _______, a__, or something else?
user: I'm interested in ___ and ____ attractions in __.
assistant: G_ has a lot to offer. Are you looking at specific ______ or re_____?
user: Yeah, I'm thinking of visiting Fre___ and M______.
assistant: Both _____o and ____his have great A__ and S_____ attractions. Let me tell you about some of them.
user: That sounds _____.

Completed conversation for example input:
assistant: Hello! What kind of attractions are you looking for? Are you interested in history, art, or something else?
user: I'm interested in Art and Scenic attractions in GA.
assistant: GA has a lot to offer. Are you looking at specific cities or regions?
user: Yeah, I'm thinking of visiting Fresno and Memphis.
assistant: Both Fresno and Memphis have great Art and Scenic attractions. Let me tell you about some of them.
user: That sounds great.

Now fill in the blanks to complete this conversation:

{masked_conv}
Completed conversation:
\end{lstlisting}

\subsubsection{In-Context Generation}

\noindent System Prompt: Same as the system prompt in Section \ref{sec:app_t1_blankfill_prompts}.

\noindent User Prompt:

\begin{lstlisting}
Here are example conversations:
{conv_list_str}

Generate 1 new similar conversation that follows the same structure.
Do not include anything other than this conversation.
Similar conversation:
\end{lstlisting}

\subsection{Prompts for BFCL}

\subsubsection{Blank Filling}

\noindent System Prompt: 

\begin{lstlisting}
You are a request completion assistant. Fill in blanks using ONLY the provided APIs and Resources. Output only the completed requests, one per line. Do not add explanations or extra text.
\end{lstlisting}

\noindent User Prompt:

\begin{lstlisting}
Fill in the blanks (underscores) to complete the user requests.
Example:
{example_context}

Input with blanks:
{example_masked}

Completed requests:
{example_completed}

Now fill in the blanks for this:
APIs: {target_classes}
Resources: {target_resources}

Input with blanks:
{masked_conv_str}

Completed requests:
\end{lstlisting}

\subsubsection{In-Context Generation}

\noindent System Prompt: 

\begin{lstlisting}
You are a test case generator. Output each request as:
Request 1: <request text>
Request 2: <request text>
\end{lstlisting}

\noindent User Prompt:

\begin{lstlisting}
{examples_str_with_APIs_resources_requests}

Generate EXACTLY {n_turns} requests for the situation below.
Requests can query information, perform operations, modify resources, or search/filter data.
Keep requests realistic. Later requests should build on earlier ones.

ONLY refer to the APIs and Resources below in the requests.
APIs: {target_classes}
Resources: {target_resources}
Requests:
\end{lstlisting}

\subsection{Prompts for ACP}

\subsubsection{Blank Filling} 
\label{sec:app_acp_blankfill_prompts}

\noindent System Prompt: 

\begin{lstlisting}
You are a question completion assistant. Fill in the blanks (underscores) to complete the question.
Use ONLY entities and terms from the provided context. Output ONLY the completed question text."
\end{lstlisting}

\noindent User Prompt: 

\begin{lstlisting}
Fill in the blanks to complete the question.
Use ONLY entities from the context.

Example 1:
Context: This is a ferry domain, where the task is to transport cars from their start to their goal locations, using a ferry. Each location is accessible by ferry from each other location. The cars can be debarked or boarded, and the ferry can carry only one car at a time. There are 3 locations and 10 cars, numbered consecutively. Currently, the ferry is at l1, with the car c2 on board. The cars are at locations as follows: c6, c3, and c0 are at l2; c4, c9, and c7 are at l0; c1, c8, and c5 are at l1.
Group: applicable_actions_bool
Answer: yes
Masked question: Is the fol___ing action appl___able in this state: deb_rk the car c2 fr_m the ferry to loc_tion l1?
Completed question: Is the following action applicable in this state: debark the car c2 from the ferry to location l1?

Example 2:
Context: This is a ferry domain, where the task is to transport cars from their start to their goal locations, using a ferry. Each location is accessible by ferry from each other location. The cars can be debarked or boarded, and the ferry can carry only one car at a time. There are 3 locations and 10 cars, numbered consecutively. Currently, the ferry is at l1 location and it is empty. The cars are at locations as follows: c9, c4, and c6 are at l0; c0, c8, c1, c7, and c2 are at l1; c3 and c5 are at l2.
Group: progression_bool
Answer: no
Masked question: Will the f_ct "The ferry is emp__" hold aft__ perf___ing the act_on "emb_rk the car c0 at loc_tion l1 on to the ferry" in the cur_ent state?
Completed question: Will the fact "The ferry is empty" hold after performing the action "embark the car c0 at location l1 on to the ferry" in the current state?

Now complete this:
Context: {target_context}
Group: {target_group}
Answer: {target_answer}
Masked question: {masked_question}
Completed question:"
\end{lstlisting}

\subsubsection{In-Context Generation}

\noindent System Prompt: 

\begin{lstlisting}
You are a test case generator for planning domain tasks. 
Given a context and examples, generate a single question that has the specified answer.
Output ONLY the question text with no preamble or label.
\end{lstlisting}

\noindent User Prompt:

\begin{lstlisting}
Examples:
{examples_str_with_context_group_answer}

Generate a question of type '{target_group}' for the context below "
such that the answer is '{target_answer}'.

Context: {target_context}

f"Question:"
\end{lstlisting}
\section{Detailed Evaluation Results}
\label{sec:app_result}

\begin{figure*}[htbp]
    \centering
\begin{subfigure}{0.48\textwidth}
         \centering
    \includegraphics[width=1\linewidth]{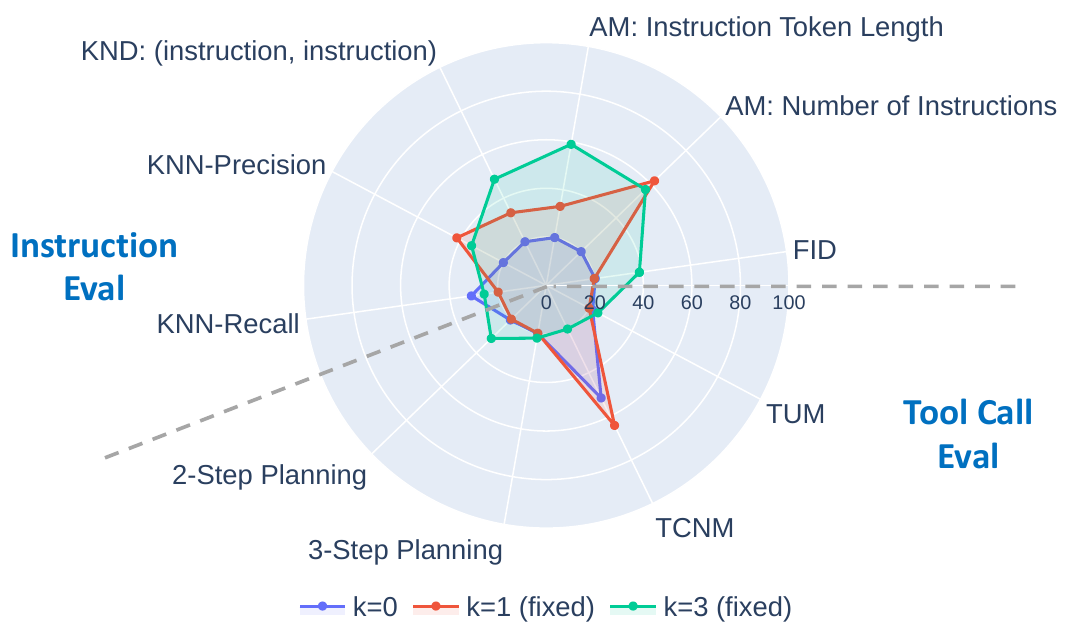}
    \caption{In-Context Generation with different $\ice{}$. Increasing $\ice{}$ does not consistently improve data utility across several metrics, including \recall{}, \precision{}, and \AM{}.\\}
    \label{fig:ice_fix_radar_bfcl}
\end{subfigure}
\begin{subfigure}{0.48\textwidth}
         \centering
    \includegraphics[width=1\linewidth]{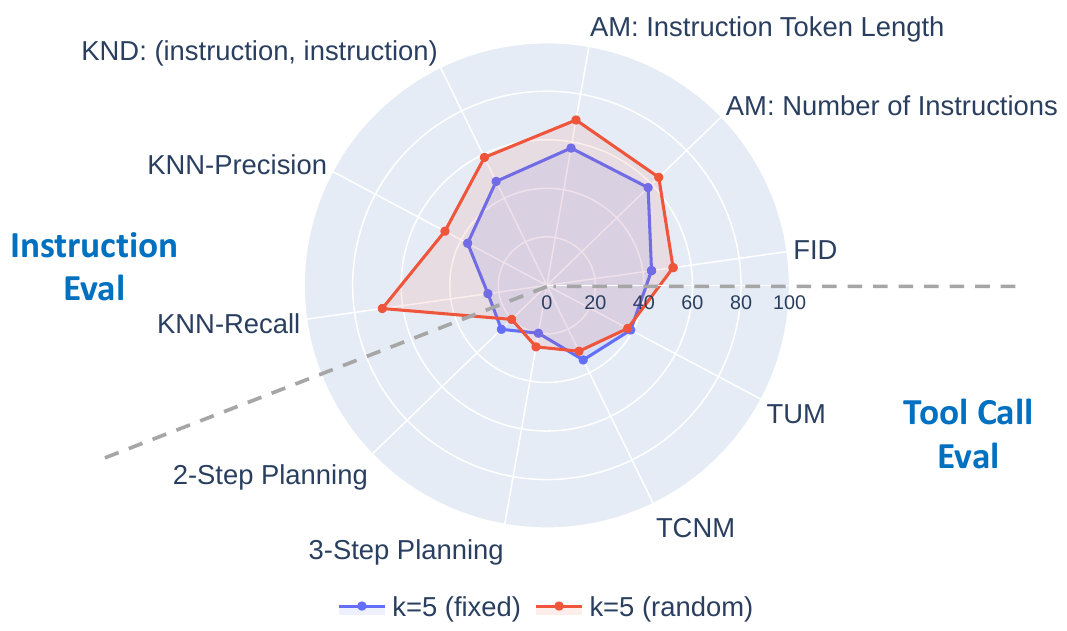}
    \caption{In-Context Generation with $\ice{}= 5$. Compared to fixed in-context examples, randomized in-context examples across generations lead to higher data utility under both instruction evaluation metrics.}
    \label{fig:ice_random_radar_bfcl}
\end{subfigure}
\vspace{0mm}
\caption{Fidelity of In-Context Generation under BFCL with fixed and randomly sampled in-context examples across generations.}
\label{fig:radar_bfcl}
\end{figure*}

The evaluation results on T1 across synthetic data generation methods are summarized in \cref{tab:result_t1}, the results on BFCL are shown in \cref{tab:result_bfcl,fig:radar_bfcl}, and the results on ACP are summarized in \cref{tab:result_acp}.

\begin{table*}[htbp]
  \caption{Synthetic data generation evaluation results on \name{} under T1.} 
\label{tab:result_t1}
  \centering
  \scalebox{0.77}{
  \begin{tabular}{llccccccccccccc}
    \toprule
            &           & \multicolumn{4}{c}{\hspace{-1mm}Instruction Eval} & \multicolumn{3}{c}{Tool Call Eval} & \multicolumn{3}{c}{\hspace{-1mm}Output Eval} & \multicolumn{2}{c}{\hspace{-1mm}Downstream Eval} \\
    \cmidrule(r){3-6}\cmidrule(r){7-9}\cmidrule(r){10-12}\cmidrule(r){13-14}
    Method & Parameter  
    & \hspace{-1mm}\KND{} $\downarrow$ 
    & \AM{} $\downarrow$ 
    & Vendi $\uparrow$ 
    & \entropy{} $\uparrow$
    & \tooluse{} $\downarrow$ 
    & \makecell{3-Step \\ Planning $\downarrow$} 
    & Vendi $\uparrow$
    & \hspace{-2mm}\makecell{KNN- \\ Precision $\uparrow$} 
    & \hspace{-2mm}\makecell{KNN- \\ Recall $\uparrow$} 
    & Vendi $\uparrow$
    & \hspace{0mm}\difficulty{} $\downarrow$ 
    & \hspace{0mm}\ranking{} $\uparrow$ \\
    \midrule

Blank Filling        
& $\blankfilling=0$   
& {0} & {0} & {9.768} & {4.260} & {0} & {0} & {2.920} & {1} & {1} & {20.085} & 0 & 1.0 \\

& $\blankfilling=0.1$ 
& 0.019 & 0.113 & {10.790} & {4.205} & 0.018 & 0.027 & {6.928} & 0.640 & 0.933 & {24.667} & 0.052 & 1.0 \\

& $\blankfilling=0.3$ 
& 0.025 & 0.096 & {12.664} & {4.246} & 0.074 & 0.134 & {9.067} & 0.587 & 0.813 & {26.354} & 0.090 & 1.0 \\

& $\blankfilling=0.5$ 
& 0.037 & 0.213 & {14.885} & {4.223} & 0.049 & 0.048 & {12.430} & 0.498 & 0.791 & {26.590} & 0.056 & 1.0 \\

& $\blankfilling=0.7$ 
& 0.035 & 0.295 & {18.029} & {3.595} & 0.022 & 0.167 & {15.159} & 0.142 & 0.796 & {41.878} & 0.147 & 0.5 \\

& $\blankfilling=0.9$ 
& 0.059 & 0.564 & {25.029} & {2.546} & 0.041 & 0.135 & {15.312} & 0.240 & 0.693 & {36.059} & 0.160 & 1.0 \\

& $\blankfilling=1$   
& 0.075 & 0.746 & {21.788} & {0.840} & 0.031 & 0.346 & {10.256} & 0.044 & 0.164 & {30.941} & 0.033 & 0.5\\
\midrule

Oversampling        
& $\oversampling=0$   
& {0.004} & {0.051} & {8.966} & {3.979} & {0} & {0} & {6.719} & {0.994} & {0.923} & {24.309} & 0.095 & 1.0 \\
& $\oversampling=0.1$ 
& 0.012 & 0.073 & {9.200} & {4.062} & 0.009 & 0.085 & {6.843} & 0.984 & 0.818 & {22.385} & 0.048 & 1.0 \\
& $\oversampling=0.3$ 
& 0.038 & 0.229 & {6.865} & {3.423} & 0.023 & 0.086 & {7.282} & 0.953 & 0.693 & {20.983} & 0.117 & 0.5 \\
& $\oversampling=0.5$ 
& 0.060 & 0.351 & {4.755} & {2.657} & 0.031 & 0.079 & {6.488} & 0.984 & 0.724 & {15.497} & 0.053 & 1.0 \\
& $\oversampling=0.7$ 
& 0.087 & 0.511 & {2.842} & {1.675} & 0.058 & 0.043 & {6.126} & 0.971 & 0.631 & {11.506} & 0.156 & 1.0 \\
& $\oversampling=0.9$ 
& 0.110 & 0.658 & {1.520} & {0.560} & 0.059 & 0.127 & {5.869} & 0.882 & 0.489 & {15.686} & 0.165 & 0.5 \\
& $\oversampling=1$   
& 0.122 & 0.750 & {1.000} & {0.000} & 0.094 & 0.238 & {5.621} & 0.992 & 0.004 & {5.986} & 0.183 & 1.0 \\
    \midrule

\multirow{7}{*}{\makecell{In-Context \\ Generation}}
& $\ice=0$   
& 0.075 & 0.746 & {21.788} & {0.840} & 0.031 & 0.346 & {10.256} & 0.044 & 0.164 & {30.941} & 0.033 & 0.5 \\

& \makecell[l]{$\ice=1$ \\ (fixed)} 
& 0.065 & 0.645 & {12.475} & {2.425} & 0.036 & 0.040 & {12.035} & 0.067 & 0.311 & {38.490} & 0.076 & 0.5 \\

& \makecell[l]{$\ice=1$ \\ (random)} 
& 0.054 & {0.444} & {15.354} & {3.466} & 0.017 & {0.008} & {12.200} & 0.027 & 0.729 & {13.259} & 0.126 & 1.0 \\

& \makecell[l]{$\ice=3$ \\ (fixed)}   
& 0.117 & 0.866 & {7.346} & {2.817} & 0.011 & 0.016 & {6.929} & 0.311 & 0.271 & {19.037} & 0.062 & 0.5 \\

& \makecell[l]{$\ice=3$ \\ (random)}   
& 0.021 & 0.523 & {10.496} & {4.168} & 0.038 & 0.034 & {7.096} & 0.062 & {0.853} & {35.427} & 0.121 & 0.5 \\

& \makecell[l]{$\ice=5$ \\ (fixed)}   
& 0.031 & 0.875 & {9.129} & {3.692} & {0.005} & 0.070 & {7.534} & 0.369 & 0.524 & {24.447} & 0.124 & 0.5 \\

& \makecell[l]{$\ice=5$ \\ (random)}   
& {0.019} & 0.460 & {10.378} & {4.305} & 0.010 & 0.085 & {6.547} & {0.471} & 0.707 & {25.269} & 0.138 & 1.0 \\
    \bottomrule
  \end{tabular}}
\end{table*}

\begin{table*}[htbp]
  \caption{Synthetic data generation evaluation results on \name{} under BFCL.} 
\label{tab:result_bfcl}
  \centering
  \scalebox{0.73}{
  \begin{tabular}{llcccccccccccc}
    \toprule
            &           & \multicolumn{5}{c}{Instruction Eval} & \multicolumn{5}{c}{Tool Call Eval} & \multicolumn{2}{c}{Downstream Eval} \\
    \cmidrule(r){3-7}\cmidrule(r){8-12}\cmidrule(r){13-14}
    Method & Parameter  
    & \hspace{-1mm}\KND{} $\downarrow$ 
    & \AM{} $\downarrow$ 
    & Vendi $\uparrow$
    & \hspace{-2mm}\makecell{KNN- \\ Precision $\uparrow$} 
    & \hspace{-2mm}\makecell{KNN- \\ Recall $\uparrow$} 
    & \tooluse{} $\downarrow$ 
    & \toolnumber{} $\downarrow$ 
    & Vendi $\uparrow$
    & \makecell{2-Step \\ Planning $\downarrow$} 
    & \makecell{3-Step \\ Planning $\downarrow$}
    & \difficulty{} $\downarrow$
    & \ranking{} $\uparrow$ \\
    \midrule

Blank Filling        
& $\blankfilling=0$   
& {0} & {0} & {22.153} & {1} & {1} & {0} & {0} & {114.256} & 0 & 0 & {0.146} & {1.0} \\

& $\blankfilling=0.1$ 
& 0.087 & 0.665 & {26.683} & 0.715 & 0.900 & 0.452 & 6.805 & {118.149} & 0.684 & 0.595 & {0.157} & {0.5} \\

& $\blankfilling=0.3$ 
& 0.079 & 1.030 & {28.295} & 0.675 & 0.905 & 0.388 & 5.445 & {109.984} & 0.684 & 0.632 & {0.163} & {0.5} \\

& $\blankfilling=0.5$ 
& 0.087 & 1.320 & {30.630} & 0.640 & 0.920 & 0.438 & 5.895 & {88.581} & 0.709 & 0.623 & {0.175} & {0.5} \\

& $\blankfilling=0.7$ 
& 0.105 & 1.470 & {32.895} & 0.430 & 0.910 & 0.656 & 16.335 & {67.985} & 0.654 & 0.567 & {0.215} & {0.5} \\

& $\blankfilling=0.9$ 
& 0.163 & 1.445 & {37.555} & 0.225 & 0.805 & 0.471 & 6.335 & {65.832} & 0.760 & 0.583 & {0.250} & {0.5} \\

& $\blankfilling=1$   
& 0.124 & 0.695 & {24.867} & 0.245 & 0.290 & 0.601 & 6.515 & {69.318} & 0.747 & 0.585 & {0.150} & {0.5} \\
\midrule

Oversampling        
& $\oversampling=0$   
& 0.020 & 0.110 & {22.058} & 1.000 & 0.875 & 0.112 & 0.285 & {52.287} & 0.156 & 0.148 & {0.005} & {1.0} \\

& $\oversampling=0.1$ 
& 0.024 & 0.130 & {23.380} & 0.995 & 0.980 & 0.150 & 0.485 & {59.895} & 0.122 & 0.085 & {0.018} & {1.0} \\

& $\oversampling=0.3$ 
& 0.033 & 0.305 & {14.432} & 0.995 & 0.965 & 0.384 & 1.395 & {30.895} & 0.261 & 0.180 & {0.022} & {0.5} \\

& $\oversampling=0.5$ 
& 0.053 & 0.525 & {8.099} & 0.995 & 0.920 & 0.583 & 2.175 & {13.555} & 0.371 & 0.258 & {0.048} & {0.5} \\

& $\oversampling=0.7$ 
& 0.080 & 0.725 & {3.804} & 0.995 & 0.910 & 0.739 & 2.985 & {5.361} & 0.466 & 0.315 & {0.073} & {0.5} \\

& $\oversampling=0.9$ 
& 0.106 & 0.955 & {1.654} & 0.995 & 0.950 & 0.850 & 3.955 & {1.824} & 0.557 & 0.368 & {0.093} & {0.5} \\

& $\oversampling=1$   
& 0.118 & 1.060 & {1.000} & 1.000 & 0.005 & 0.912 & 4.290 & {1.000} & 0.627 & 0.411 & {0.108} & {0.5} \\
    \midrule

\multirow{7}{*}{\makecell{In-Context \\ Generation}}
& $\ice=0$   
& 0.124 & 0.695 & {24.867} & 0.245 & 0.290 & 0.601 & 6.515 & {69.318} & 0.747 & 0.585 & {0.150} & {0.5} \\

& \makecell[l]{$\ice=1$ \\ (fixed)} 
& 0.103 & 0.330 & {15.082} & 0.450 & 0.175 & 0.613 & 4.820 & {66.247} & 0.752 & 0.586 & {0.138} & {0.5} \\

& \makecell[l]{$\ice=1$ \\ (random)} 
& 0.129 & 0.570 & {29.312} & 0.370 & 0.760 & 0.571 & 5.045 & {53.051} & 0.698 & 0.596 & {0.182} & {0.5} \\

& \makecell[l]{$\ice=3$ \\ (fixed)}   
& 0.079 & 0.375 & {20.832} & 0.385 & 0.235 & 0.582 & 10.770 & {91.796} & 0.644 & 0.571 & {0.135} & {0.5} \\

& \makecell[l]{$\ice=3$ \\ (random)}   
& 0.115 & 0.515 & {29.471} & 0.500 & 0.765 & 0.557 & 7.430 & {65.258} & 0.699 & 0.574 & {0.162} & {0.5} \\

& \makecell[l]{$\ice=5$ \\ (fixed)}   
& 0.081 & 0.365 & {24.618} & 0.405 & 0.330 & 0.457 & 5.360 & {94.033} & 0.691 & 0.621 & {0.122} & {0.5} \\

& \makecell[l]{$\ice=5$ \\ (random)}   
& 0.064 & 0.312 & {27.739} & 0.505 & 0.720 & 0.467 & 5.685 & {56.282} & 0.745 & 0.577 & {0.152} & {0.5} \\
    \bottomrule
  \end{tabular}}
\end{table*}

\begin{figure*}[htbp]
    \centering
\begin{subfigure}{0.48\textwidth}
         \centering
    \includegraphics[width=1\linewidth]{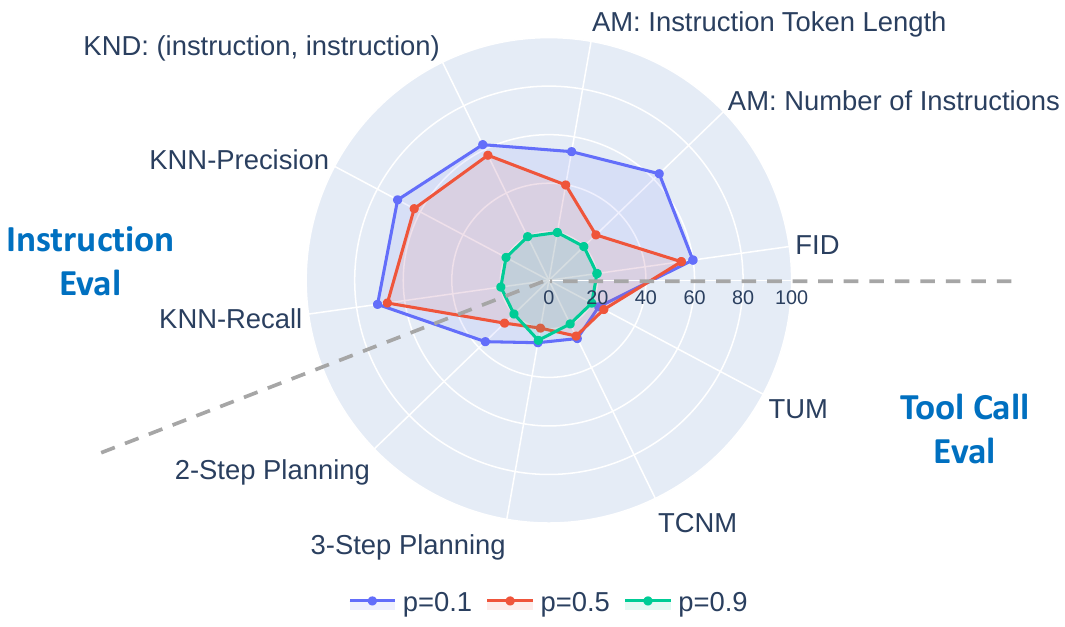}
    \vspace{0mm}
    \caption{Blank Filling. As $\blankfilling{}$ increases, data utility degrades across nearly all metrics on fiedelity.}
    \label{fig:blankfill_radar_bfcl}
\end{subfigure}
\begin{subfigure}{0.48\textwidth}
         \centering
    \includegraphics[width=1\linewidth]{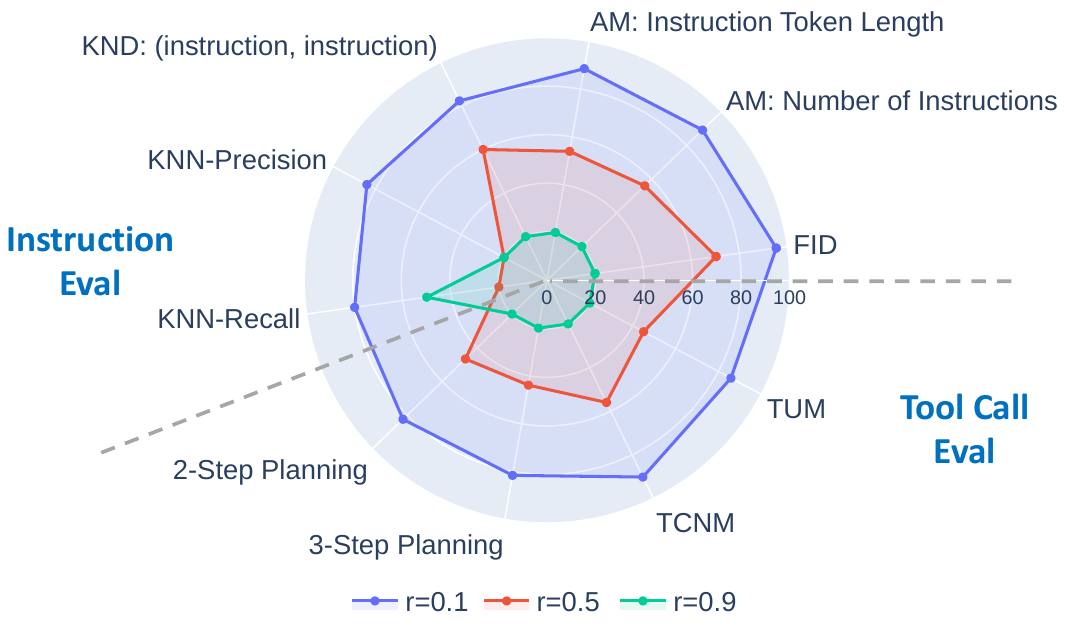}
    \vspace{0mm}
    \caption{Oversampling. As $\oversampling{}$ increases, data utility degrades across nearly all metrics on fidelity.}
    \label{fig:oversample_radar_bfcl}
\end{subfigure}
\vspace{0mm}
\caption{Fidelity of Blank Filling and Oversampling on the BFCL dataset.}
\label{fig:radar_bfcl_1}
\end{figure*}

\begin{table*}[htbp]
  \caption{Synthetic data generation evaluation results on \name{} under ACP.} 
\label{tab:result_acp}
  \centering
  \scalebox{0.9}{
  \begin{tabular}{llccccccccc}
    \toprule
            &           & \multicolumn{4}{c}{Instruction Eval} & \multicolumn{3}{c}{Output Eval} & \multicolumn{2}{c}{Downstream Eval} \\
    \cmidrule(r){3-6}\cmidrule(r){7-9}\cmidrule(r){10-11}
    Method & Parameter  
    & \hspace{-1mm}\KND{} $\downarrow$ 
    & \AM{} $\downarrow$
    & FID $\downarrow$ 
    & Vendi $\uparrow$
    & \hspace{-2mm}\makecell{KNN- \\ Precision $\uparrow$} 
    & \hspace{-2mm}\makecell{KNN- \\ Recall $\uparrow$} 
    & \successrate $\uparrow$
    & \difficulty{} $\downarrow$
    & \ranking{} $\uparrow$ \\
    \midrule

Blank Filling        
& $\blankfilling=0$   
& {0.011} & {3.319} & {0.000} & {10.822} & {1.000} & {1.000} & {0.627} & {0.042} & {0.5} \\

& $\blankfilling=0.1$ 
& {0.049} & {10.981} & {0.000} & {10.822} & {1.000} & {1.000} & {0.638} & {0.088} & {0.5} \\

& $\blankfilling=0.3$ 
& {0.081} & {8.650} & {0.000} & {10.822} & {1.000} & {1.000} & {0.662} & {0.106} & {0.5} \\

& $\blankfilling=0.5$ 
& {0.150} & {9.142} & {0.000} & {10.822} & {1.000} & {1.000} & {0.642} & {0.101} & {0} \\

& $\blankfilling=0.7$ 
& {0.157} & {13.346} & {0.000} & {10.822} & {1.000} & {1.000} & {0.619} & {0.103} & {0.5} \\

& $\blankfilling=0.9$ 
& {0.176} & {18.992} & {0.000} & {10.822} & {0.996} & {1.000} & {0.600} & {0.126} & {0} \\

& $\blankfilling=1$   
& {0.203} & {21.746} & {3.981} & {10.447} & {0.996} & {0.985} & {0.538} & {0.142} & {0.5}\\
\midrule

Oversampling        
& $\oversampling=0$   
& {0.007} & {10.769} & {5.195} & {10.386} & {0.988} & {0.988} & {0.827} & {0.044} & {1.0} \\

& $\oversampling=0.1$ 
& {0.018} & {10.327} & {5.235} & {10.506} & {0.996} & {1.000} & {0.731} & {0.036} & {1.0} \\

& $\oversampling=0.3$ 
& {0.060} & {31.735} & {33.454} & {8.430} & {0.992} & {0.985} & {0.638} & {0.133} & {1.0} \\

& $\oversampling=0.5$ 
& {0.094} & {52.988} & {83.330} & {5.450} & {0.981} & {0.992} & {0.565} & {0.206} & {1.0} \\

& $\oversampling=0.7$ 
& {0.135} & {80.908} & {150.814} & {3.177} & {0.977} & {0.962} & {0.438} & {0.301} & {1.0} \\

& $\oversampling=0.9$ 
& {0.170} & {99.827} & {263.393} & {1.582} & {0.969} & {0.962} & {0.323} & {0.403} & {1.0} \\

& $\oversampling=1$   
& {0.188} & {114.935} & {382.144} & {1.000} & {0.969} & {0.019} & {0.242} & {0.455} & {1.0} \\
    \midrule

\multirow{7}{*}{\makecell{In-Context \\ Generation}}
& $\ice=0$   
& {0.203} & {21.746} & {3.981} & {10.447} & {0.996} & {0.985} & {0.538} & {0.142} & {0.5} \\

& \makecell[l]{$\ice=1$ \\ (fixed)} 
& {0.184} & {16.085} & {3.981} & {10.447} & {0.981} & {0.985} & {0.515} & {0.191} & {0.5} \\

& \makecell[l]{$\ice=1$ \\ (random)} 
& {0.213} & {30.408} & {2.736} & {10.505} & {0.992} & {0.977} & {0.538} & {0.160} & {0.5} \\

& \makecell[l]{$\ice=3$ \\ (fixed)}   
& {0.225} & {17.681} & {4.096} & {10.435} & {0.992} & {0.985} & {0.558} & {0.142} & {0} \\

& \makecell[l]{$\ice=3$ \\ (random)}   
& {0.186} & {15.085} & {3.760} & {10.507} & {0.996} & {1.000} & {0.581} & {0.127} & {0.5} \\

& \makecell[l]{$\ice=5$ \\ (fixed)}   
& {0.146} & {15.181} & {4.760} & {10.362} & {0.988} & {0.973} & {0.612} & {0.109} & {0.5} \\

& \makecell[l]{$\ice=5$ \\ (random)}   
& {0.130} & {11.792} & {2.464} & {10.773} & {0.992} & {0.977} & {0.550} & {0.150} & {0.5} \\
    \bottomrule
  \end{tabular}}
\end{table*}
\section{\revise{Evaluation Results of Nvidia NeMo}}
\label{sec:app_nemo}

To evaluate a more realistic synthetic dataset, we use NVIDIA NeMo, an industry-standard synthetic data tool (including for testing agents), to generate synthetic datasets using three backbone models, \texttt{mistral-small-24b-instruct}, \texttt{nvidia-nemotron-nano-9b-v2} and \texttt{Llama3.1-8B-Instruct}, and evaluate them with SynAE on the T1 dataset. Our results in \cref{tab:nemo-backbones} are consistent with prior findings that the three models have similar performance on standard LLM benchmarks: SynAE reports similar fidelity scores for all. However, SynAE also shows that Nemotron and Llama exhibits greater diversity than Mistral. We use this observation to explore the distribution of the \emph{attraction type} attribute in \cref{tab:attraction-type-distribution}, where Nemotron and Llama show greater variability than Mistral. This illustrates that SynAE can uncover non-obvious properties of synthetic agent trajectories.

We also observe from \cref{tab:nemo-backbones} that, for both backbone models, increasing the generation temperature lowers KNN-Precision and increases KNN-Recall (coverage). This is intuitive, since higher temperatures introduce more randomness into generation and thus increase coverage.

To show that SynAE captures real evaluation outcomes, we compare agent performance on Nemotron datasets generated with temperatures 0.1 and 0.5 in \cref{tab:agent-performance}. Relative to $tmp=0.1$, agent performance on Nemotron $tmp=0.5$ deviates more from the real dataset, which is reflected by the larger TDD and lower RD in \cref{tab:nemo-backbones}. This indicates larger performance discrepancies and less consistent rankings.

\begin{table*}[htbp]
    \centering
    \caption{NeMo evaluation results on SynAE under T1.}
    \label{tab:nemo-backbones}
    \scalebox{0.7}{
    \begin{tabular}{llcccccccccc}
        \toprule
        \multirow{2}{*}{Model} & \multirow{2}{*}{Temperature}
        & \multicolumn{1}{c}{Validity}
        & \multicolumn{2}{c}{\shortstack{Fidelity:\\Instruction Eval}}
        & \multicolumn{1}{c}{\shortstack{Fidelity:\\Tool Call Eval}}
        & \multicolumn{2}{c}{\shortstack{Fidelity:\\Output Eval}}
        & \multicolumn{2}{c}{\shortstack{Fidelity:\\Downstream Eval}}
        & \multicolumn{2}{c}{Diversity} \\
        \cmidrule(lr){3-3}
        \cmidrule(lr){4-5}
        \cmidrule(lr){6-6}
        \cmidrule(lr){7-8}
        \cmidrule(lr){9-10}
        \cmidrule(lr){11-12}
        &
        & \shortstack{VR:\\ output} $\uparrow$
        & \shortstack{KND:\\ (instruction, response) $\downarrow$}
        & \shortstack{AM:\\ city} $\downarrow$
        & \shortstack{3-step\\planning $\downarrow$}
        & \shortstack{KNN-\\precision} $\uparrow$
        & \shortstack{KNN-\\recall} $\uparrow$
        & TDD $\downarrow$
        & RD $\uparrow$
        & Vendi score $\uparrow$
        & AD $\uparrow$ \\
        \midrule
        \multirow{5}{*}{Mistral}
        & $tmp=0.1$ & 0.798 & 0.131 & 0.828 & 0.540 & 0.382 & 0.089 & 0.110 & 1.0 & 4.692 & 2.449 \\
        & $tmp=0.3$ & 0.831 & 0.149 & 0.869 & 0.522 & 0.302 & 0.067 & 0.044 & 1.0 & 4.759 & 2.592 \\
        & $tmp=0.5$ & 0.898 & 0.135 & 0.805 & 0.626 & 0.222 & 0.338 & 0.011 & 1.0 & 5.472 & 2.899 \\
        & $tmp=0.7$ & 0.811 & 0.143 & 0.932 & 0.519 & 0.227 & 0.369 & 0.096 & 1.0 & 6.305 & 2.718 \\
        & $tmp=0.9$ & 0.893 & 0.113 & 0.715 & 0.492 & 0.013 & 0.533 & 0.274 & 1.0 & 7.472 & 3.077 \\
        \midrule
        \multirow{5}{*}{Nemotron}
        & $tmp=0.1$ & 0.764 & 0.112 & 0.903 & 0.551 & 0.209 & 0.271 & 0.129 & 1.0 & 5.145 & 2.347 \\
        & $tmp=0.3$ & 0.787 & 0.117 & 0.843 & 0.404 & 0.244 & 0.298 & 0.141 & 1.0 & 6.013 & 2.676 \\
        & $tmp=0.5$ & 0.820 & 0.115 & 0.806 & 0.482 & 0.237 & 0.302 & 0.323 & 0.5 & 7.619 & 3.125 \\
        & $tmp=0.7$ & 0.778 & 0.106 & 0.819 & 0.520 & 0.031 & 0.902 & 0.203 & 0.5 & 8.246 & 3.160 \\
        & $tmp=0.9$ & 0.856 & 0.104 & 0.780 & 0.476 & 0.049 & 0.667 & 0.397 & 1.0 & 9.524 & 3.487 \\
                \midrule
        \multirow{5}{*}{Llama}
        & $tmp=0.1$ & 0.809 & 0.123 & 0.789 & 0.729 & 0.120 & 0.662 & 0.145 & 0.5 & 7.480 & 3.184 \\
        & $tmp=0.3$ & 0.832 & 0.142 & 0.765 & 0.817 & 0.037 & 0.511 & 0.161 & 1.0 & 8.299 & 3.187 \\
        & $tmp=0.5$ & 0.881 & 0.129 & 0.795 & 0.571 & 0.259 & 0.160 & 0.135 & 0.5 & 9.236 & 3.257 \\
        & $tmp=0.7$ & 0.804 & 0.138 & 0.798 & 0.497 & 0.286 & 0.236 & 0.157 & 0.5 & 8.788 & 3.408 \\
        & $tmp=0.9$ & 0.859 & 0.135 & 0.770 & 0.660 & 0.100 & 0.427 & 0.154 & 0.5 & 9.397 & 3.481 \\
        \bottomrule
    \end{tabular}
    }
\end{table*}

\begin{table*}[htbp]
    \centering
    \caption{Distribution of attraction types.}
    \label{tab:attraction-type-distribution}
    \begin{tabular}{lccccc}
        \toprule
        Model & Art & Historical & Cultural & Scenic & Other \\
        \midrule
        \texttt{mistral-small-24b-instruct} ($tmp=0.7$)   & 35.73\% & 20.62\% &  9.08\% &  5.76\% & 28.80\% \\
        \texttt{nvidia-nemotron-nano-9b-v2} ($tmp=0.7$)  & 25.76\% & 23.80\% & 13.21\% & 12.77\% & 24.46\% \\
        \texttt{Llama3.1-8B-Instruct} ($tmp=0.7$)  & 25.47\% & 24.07\% & 12.84\% & 11.01\% & 26.61\% \\
        \bottomrule
    \end{tabular}
\end{table*}

\begin{table*}[htbp]
    \centering
    \caption{Agent performance on real and synthetic datasets.}
    \label{tab:agent-performance}
    \begin{tabular}{lccc}
        \toprule
        Dataset & Agent 1 & Agent 2 & Agent 3 \\
        \midrule
        real dataset         & 1.0000 $\pm$ 0 & 0.6356 $\pm$ 0.0098 & 0.9600 $\pm$ 0.0067 \\
        \texttt{nvidia-nemotron-nano-9b-v2} ($tmp=0.1$) & 1.0000 $\pm$ 0 & 0.5644 $\pm$ 0.0102 & 0.6444 $\pm$ 0.0049 \\
        \texttt{nvidia-nemotron-nano-9b-v2} ($tmp=0.5$) & 1.0000 $\pm$ 0 & 0.3200 $\pm$ 0.0131 & 0.2967 $\pm$ 0.0081 \\
        \bottomrule
    \end{tabular}
\end{table*}
\section{\revise{No single baseline metric can fully characterize synthetic data performance}}
\label{app:baseline}

We report several simple baselines for evaluating synthetic instructions and responses generated by In-Context Generation with different numbers of in-context examples, $k$. Specifically, we consider vocabulary-overlap F1, total variation (TV) distance between unigram distributions, TV distance on length distributions, Fr\'echet Inception Distance (FID). The results are shown in \cref{tab:icg_simple_baselines}.

\begin{table}[htbp]
\centering
\caption{Simple baseline metrics for synthetic instructions and responses generated by In-Context Generation with different numbers of in-context examples $k$.}
\vspace{4mm}
\setlength{\tabcolsep}{6pt}
\begin{tabular}{lcccc}
\toprule
Metric & $k=0$ & $k=1$ & $k=3$ & $k=5$ \\
\midrule
Vocabulary overlap (F1) $\uparrow$ & 0.3856 & 0.4788 & 0.6030 & 0.6809 \\
Unigram distribution (TV) $\downarrow$ & 0.6132 & 0.5281 & 0.3891 & 0.3111 \\
Length distribution $\downarrow$ & 9.2444 & 8.7733 & 8.7422 & 8.7022 \\
FID $\downarrow$ & 204.0479 & 166.2299 & 131.6828 & 107.5157 \\
\bottomrule
\end{tabular}
\label{tab:icg_simple_baselines}
\end{table}

Taken in isolation, these baselines may suggest that increasing $k$ monotonically improves synthetic data quality, since lexical overlap increases while several distributional distances decrease. However, this conclusion is misleading. As shown in \cref{fig:ice_fix_radar}, using more in-context examples does not necessarily improve overall generation quality. This is consistent with prior work showing that in-context learning is highly sensitive to the choice of demonstrations, and that adding more demonstrations does not uniformly improve performance across tasks \citep{liu2022makes,zou2025many}. More broadly, these simple baselines are useful as sanity checks, but they mainly capture surface-form similarity or coarse marginal statistics, and therefore can miss higher-order aspects of synthetic data quality such as validity and diversity.
\section{Attribute Distribution of the Skewed and Augmented Datasets}
\label{app:case_study}

We report the attraction-type distributions of the skewed and augmented datasets in \cref{tab:case_study_attraction_type}. The skewed dataset contains four under-represented attraction types, defined as those with proportions below $5\%$, whereas the augmented datasets produced by Relabeling or NVIDIA NeMo contain no under-represented attraction types.

\begin{table*}[htbp]
    \centering
    \caption{Distribution of attraction types. \red{Red} indicates under-represented attraction types.}
    \label{tab:case_study_attraction_type}
    \resizebox{\textwidth}{!}{
    \begin{tabular}{lccccccccc}
        \toprule
        Dataset & Art & Historical & Cultural & Scenic & Touristy & Culinary & Guided & Social & Sporting \\
        \midrule
        Skewed Real Data        & 16.67\% & 15.56\% &  \red{2.22\%} & 21.11\% & 17.78\% &  \red{4.44\%} &  \red{4.96\%} & 12.82\% &  \red{4.44\%} \\
        Relabeling              & 12.60\% & 11.02\% &  9.45\% & 16.54\% & 12.60\% & 11.02\% &  9.45\% &  8.66\% &  8.66\% \\
        NVIDIA NeMo (Llama)     & 27.35\% & 13.76\% & 11.59\% & 12.13\% &  9.42\% &  7.24\% &  6.70\% &  6.70\% &  5.07\% \\
        NVIDIA NeMo (Nemotron)  & 19.56\% & 14.92\% & 12.83\% & 11.28\% & 11.80\% &  9.74\% &  8.19\% &  6.13\% &  5.55\% \\
        NVIDIA NeMo (GPT)       & 17.74\% & 14.07\% &  12.74\% & 12.74\% & 12.07\% &  9.40\% &  6.07\% &  7.40\% &  7.74\% \\
        \bottomrule
    \end{tabular}
    }
\end{table*}

\end{document}